# A comparative study on the accuracy & repeatability of mobile robotic platforms for the delivery of precision NDE measurement

SeyedMohammadAmin Nabi Pour, S. Gareth Pierce, Randika Vithanage, Ehsan Mohseni, David Carswell, Matthew Shields

## Abstract

Mobile robotic platforms offer a flexible alternative to fixed manipulators for non destructive evaluation (NDE) of large aerospace structures, but their base-positioning accuracy and how that accuracy should inform deployment has not been assessed under a common, externally referenced protocol. This work presents a laser tracker based evaluation workflow (ground truth ~6 µm) that measures the static and segmented trajectory positioning accuracy of five commercial mobile platforms (KUKA KMP-1500, KUKA KMR, MiR250, Boston Dynamics Spot, Clearpath Husky) under a common protocol. A coupled multi corner calibration recovers the laser to robot transformation and reflector offsets; ordinary least squares over all poses is used, with robust estimation retained only as a blunder check.

Static positioning accuracy ranged from a median of 8.2 mm (KMP-1500) to 63.5 mm (Spot), with the wheel odometry only Husky uncalibratable. Dynamic path following was characterised by cross track error; The component insensitive to temporal alignment which ranged from 6.9 mm (KMP-1500) to 112.1 mm (Spot). Both accuracy and calibratability tracked localisation capability, from the newest LiDAR SLAM platform to map free visual odometry. No configuration meets the ±0.2–1.0 mm aerospace NDE tolerance from the base alone; the results are framed as a design input that sizes the supplementary sensing each platform requires roughly one order of magnitude for the best platform, nearly two for the worst providing a reproducible basis for platform selection rather than a feasibility claim.

## 1. Introduction

Non-destructive evaluation (NDE) is a multidisciplinary field that underpins many manufacturing and in service quality assurance checks; it is particularly widely deployed in high value structures such as aircraft, pressure vessels, critical infrastructure, nuclear, defence, renewable energy, etc [1], [2].

The aerospace sector relies heavily on NDE to ensure the safety of primary components, including flight control surfaces, fuselages, and propulsion systems, and it is deployed both during initial manufacture and throughout life maintenance, overhaul, and repair (MRO) activities. Strict standards govern the sector to ensure the structural integrity and safety of aircraft components. These standards often involve cumbersome, time consuming procedures to verify each part, especially when dealing with large structures many metres in size. [3]

Robotic systems have been deployed to automate and accelerate NDE data acquisition. Industrial robots equipped with ultrasonic transducers have demonstrated effectiveness for automated inspection of complex curved components such as aero engine blades, achieving high accuracy in both defect detection and thickness measurement [4]. These hardware systems are now being augmented with machine learning techniques to accelerate data interpretation and improve defect detection accuracy [6]. Traditional fixed robotic manipulators are widely used in NDE applications due to their high accuracy and repeatability (typically down to 0.1 mm), guaranteeing excellent spatial registration of NDE data with component geometry. However, such manipulators are inflexible in their programming and typically configured in fixed cells that allow scanning of only a single or limited number of component geometries. Recent advancements in collaborative and mobile robotic platforms have generated interest in using such systems to provide a more flexible approach to NDE measurements, enabling deployment with limited prior knowledge of the component to be scanned [7]. The integration of mobile manipulators (Figure 1.1) into NDE processes could significantly improve flexibility by automating inspection tasks and enabling access to complex geometries, potentially accelerating non-destructive testing procedures while maintaining compliance with stringent aerospace regulations. The authors have recently demonstrated [8] an approach capable of delivering multiple NDE measurement modalities to components of unknown surface geometry using mobile robotic systems.

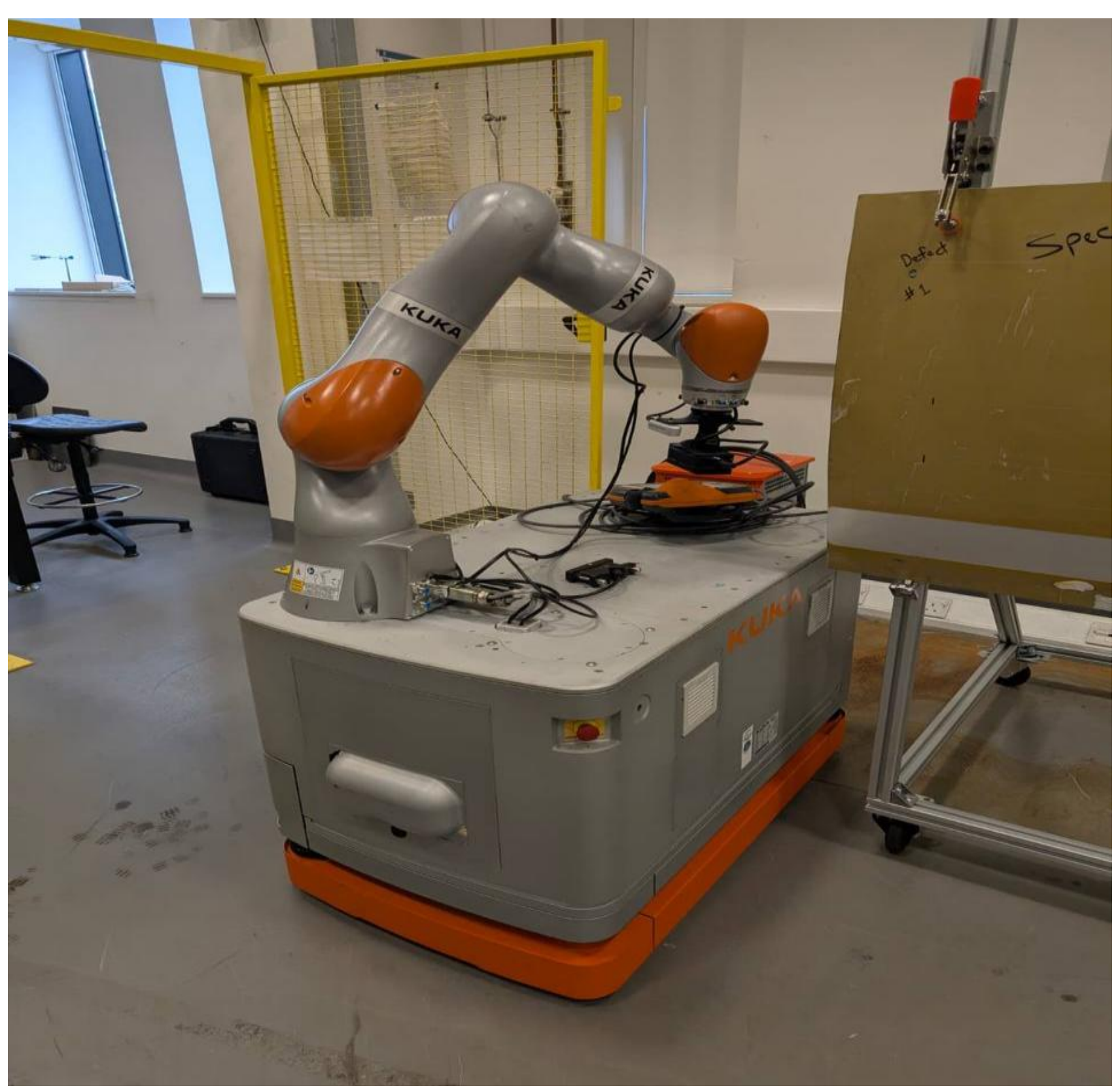

Fig 1.1 : KUKA KMR is a mobile manipulator platform.

This approach, however, adds complexity to operations, with one of the primary challenges being the uncertainty of the mobile platform's position within the workspace. This is critical because the positioning of the manipulator depends on the base, which, in this case, is the mobile platform [9].

Mobile manipulators offer promising solutions, especially for tasks involving large structures that traditionally required bulky, stationary industrial robots. By using smaller, more agile mobile manipulators, industries can implement more flexible and scalable manufacturing processes [10] This study therefore concentrates on the assessment of accuracy and repeatability of multiple popular robotic mobile base units, and assesses the suitability for advanced NDE inspection applications in a more flexible factory environment.

Currently, the International Organisation for Standardisation (ISO) provides documentation outlining standard procedures and requirements for robotic NDT inspection. However, these standards do not specifically address mobile manipulators or how the accuracy of a mobile platform may impact the overall precision of the manipulator it carries [11], [12]. Furthermore, existing ISO standards for mobile service robots primarily focus on obstacle detection and avoidance. They do not extend to aspects such as positional accuracy or repeatability, which are critical in high precision tasks like NDT [13]. This study, therefore, serves to fill this knowledge gap.

To address the lack of standard procedures and reliable data on mobile platform accuracy, this study developed and conducted a systematic evaluation of five mobile platforms (KUKA KMP1500 (Figure 1.5), KUKA KMR (Figure 1.4) , MiR250 (Figure 1.2) , Boston Dynamics Spot Dog (Figure 1.3), and Clearpath Robotics Husky (Figure 1.6). The goal was to measure both their static and dynamic accuracy and repeatability under identical experimental conditions to gain a comparative understanding of their potential performance when used as mobile delivery platforms for NDE or manipulators for associated manufacturing.

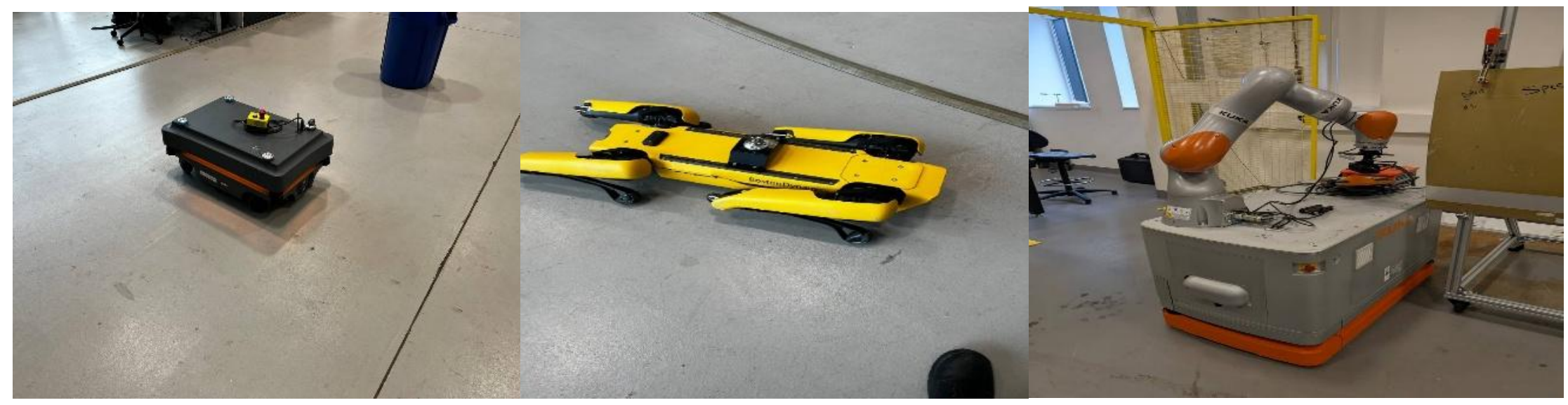

Fig 1.2: MiR250. Fig 1.3: Boston Dynamic Spot Dog. Fig 1.4: KUKA KMR platform.

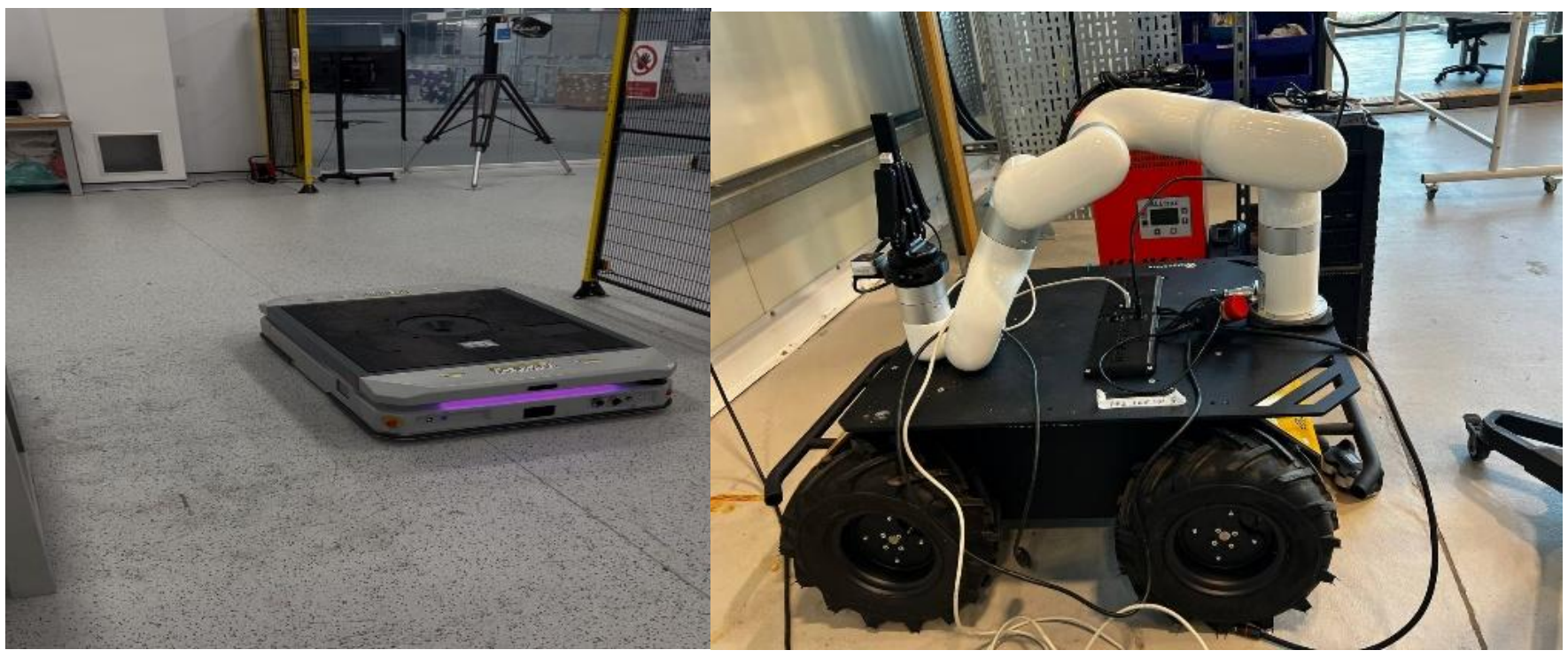

Fig 1.5: KUKA KMP-1500. Fig 1.6: Clearpath Husky.

The primary objective of this study is therefore to quantify the accuracy and repeatability of multiple mobile platforms operating in controlled conditions and to compare their relative performance. This evaluation provides essential baseline data for future deployment of these platforms with manipulator arms, enabling assessment of combined mobile manipulator accuracy for NDT applications.

The key contributions of this work are:

1. A standardised, externally referenced experimental workflow for quantifying mobile platform base positioning accuracy both static and along segmented trajectories that is reproducible across heterogeneous commercial platforms and control stacks.
2. A comparative characterisation of five commercial platforms under identical conditions, reporting the full error distribution (including the tail of poor placements) rather than a single flattered mean.
3. A localisation capability cascade linking each platform's positioning accuracy and its very calibratability to its onboard localisation, and a framework that uses this to size the supplementary sensing required for NDE deployment.

The study characterises the mobile base as the precursor to a mobile manipulator; it does not claim to measure end effector or probe accuracy, which would additionally depend on the manipulator and end effector calibration.

## 2. Background

Robots can generally be divided into two categories: fixed robots, primarily industrial robots, and mobile platforms [14]. Industrial robots are mostly used for repetitive tasks but often lack the flexibility required for varying production processes [15]. Recent advancements in software and control have improved the accuracy of industrial robots, enabling their use in a wide range of industries, including aerospace and, more specifically, non-destructive automation testing [16].

In recent years, Power and Force Limiting (PFL) robots have been introduced to enhance the flexibility of traditional industrial robots. PFL robots are safer to operate around humans, making them more attractive to industries as an alternative to conventional industrial robots [17]. However, PFL robots are typically stationary, which limits their reach and flexibility for diverse tasks [18].

Mobile robot platforms have been utilised in industries since the 1950s, and their applications has grown significantly over time. These platforms are widely used for transporting goods and items between locations [19]. Another important category of robots is mobile manipulators, which combine a mobile platform with a fixed or collaborative robotic arm. These systems are developed to integrate the flexibility of mobile platforms with the precision and capabilities of industrial or collaborative robots. Their adoption in industrial and manufacturing environments brings unique challenges, including safe operation around humans, reliable navigation in dynamic spaces, and innovative methods of material handling and task execution. These challenges have been the focus of ongoing research.

All five platforms (KUKA KMR, KUKA KMP-1500, Boston Dynamics Spot, MiR250, Clearpath Husky) are capable of integrating with onboard robotic arms such as the Universal Robot, KUKA LBR iiwa, or the Spot Arm, making them highly suitable for industrial automation and inspection tasks. Moreover, four of these platforms (MiR250, Clearpath

Husky, KUKA KMR, and Boston Dynamics Spot) provide manufacturer supported ROS (Robot Operating System) interfaces, whilst the KUKA KMP-1500 utilises its native control software. The ROS enabled platforms benefit from standardised communication protocols and a flexible, scalable framework for real time control and automation [20], [21], [22], [23] [24].

### 2.1 Navigation and Path Planning Methods

One of the main aspects of designing and developing any mobile robot is navigation. The robot must be capable of navigating in both known and unknown environments. This includes determining its location, understanding its configuration, controlling its actuators to achieve goals, and interpreting sensor readings to avoid obstacles [5]. Mobile platforms can navigate and localise themselves using two primary methods. The first relies solely on odometry, where data is continuously collected from wheels or legs, depending on the robot's configuration. The second combines odometry with additional sensors, such as LiDAR or vision cameras, enabling the robot to detect environmental features or landmarks that can be matched to known maps or reference points. Using only odometry accumulates an integrated path error (due to wheel slip, etc.), which increases over time. External sensor measurements can be fused with odometry to bound these drift errors and improve navigation precision through techniques such as SLAM (Simultaneous Localisation and Mapping) or marker based localisation [14].

One categorisation of mobile platform navigation is based on the algorithms used, which can be divided into classical and heuristic methods. Classical methods include Artificial Potential Fields [25], Dijkstra's Algorithm [26], Cell Decomposition [27], and Probabilistic Road Maps [28]. Heuristic methods include Fuzzy Logic [29], Neural Networks [30], Genetic Algorithms [31], Particle Swarm Optimisation [32], Artificial Bee Colony [33], and Cuckoo Search Optimisation [34]. Some studies also propose hybrid methods by combining two approaches, classical and heuristic, to address their individual limitations [35]. Additionally, other research divides these methods into four categories: classical methods, heuristic methods, bio inspired methods, and AI based methods, allowing for more independent investigation [36].

Path planning techniques are often divided into two main categories: global path planning and local path planning. Global path planning focuses on creating a path from a start point to a goal while avoiding obstacles, typically fixing the path before the robot starts moving. Local path planning, on the other hand, continuously makes adjustments based on changes in the environment, such as dynamic obstacles, while following the global path. Classical approaches are generally preferred for global path planning due to their fast computation and low memory requirements. In contrast, AI based methods are better suited for local path planning, as they handle environmental uncertainties more effectively [37]

As it mentioned earlier path planning and localisation is a challenging task in mobile platforms. Researchers continue to develop various methods to enable mobile platforms to navigate different environments [38]. These navigation and localisation techniques have their strengths and weaknesses, performing differently in known and unknown environments with static and dynamic obstacles. However, all these methods aim to guide the mobile platform from a starting point to a destination in the shortest, fastest, and safest manner without collisions [36]. While much focus has been placed on obstacle avoidance, there is limited research on the accuracy and repeatability of these platforms in following their intended paths and reaching their destinations, which can significantly impact industrial applications [15].

### 2.2 Localisation and Mapping Technologies and Performance

Owing to speed of computation and data streaming, most modern mobile platforms use SLAM or localisation and mapping. SLAM allows robots to simultaneously build a map of an unknown environment and localise themselves within it using odometry and sensor data. SLAM methods are typically divided into two main types: visual SLAM, which uses dense visual data for mapping, and laser SLAM, which employs LiDAR scanners to continuously monitor the robot's surroundings [39].

SLAM methods, however, are affected by environmental factors such as area features, lighting conditions, and surface reflectivity. These dependencies can introduce noise into sensor readings, impacting mapping and localisation accuracy. While significant advancements have been made to address these limitations, further research is needed to enhance SLAM systems [40].

To overcome these challenges, many new methods have been introduced in recent years. One effective approach is sensor fusion, which combines multiple sensors to improve accuracy. For instance, Radar Inertial Navigation SLAM fuses radar and IMU data for precise localisation and mapping, while LiDAR-Visual-IMU SLAM integrates camera based feature extraction with laser and IMU data to achieve higher accuracy [41].

Another way to enhance SLAM accuracy is through optimisation methods, such as fusion SLAM [42]mentioned above, or integrating SLAM with machine learning techniques. AI and deep learning algorithms significantly improve SLAM's capabilities by extracting robust features and providing a semantic level understanding of the mapped environment [43]. However, these methods rely heavily on large amounts of annotated training data and may require considerable computational resources. Nevertheless, they hold promise for enabling smarter mapping and localisation in future robots [41].

### 2.3 Research Gaps and Limitations in Mobile Robot Accuracy Assessment

The accuracy limitations observed in this study align with fundamental challenges inherent to mobile robotic platforms identified in recent literature. Localisation remains a primary constraint on mobile platform positioning, as techniques including LiDAR, vision systems, GPS, UWB, and inertial navigation all rely on odometry measurements whose accuracy is inherently bounded by the resolution and precision of integrated sensors [44]. Beyond sensor limitations, mechanical factors significantly degrade positioning performance. Wheel slippage represents a particularly problematic error source, as propulsion forces depend on frictional contact with the ground surface a relationship subject to considerable uncertainty due to the mobile base's large inertia and variable ground conditions. This wheel ground contact uncertainty means that mobile platform kinematic accuracy is fundamentally lower than that achievable with fixed manipulators. These challenges are compounded in unstructured or dynamic industrial environments where obstacles, moving personnel, and uneven terrain introduce additional disturbances including wheel slip, slide, and unpredictable surface conditions [45].

Despite significant progress in localisation, SLAM, navigation and path planning for mobile platforms, much of the research remains confined to simulations rather than real world environments. Testing algorithms solely in simulations introduces uncertainties that may not translate well to practical applications [37]. Furthermore, there is no standard or reference method for evaluating how different commercially available robots perform in SLAM. Manufacturers also often provide limited data, making it challenging to assess and compare the performance of these systems [15].

Evaluating mobile platform localisation accuracy presents significant challenges due to the need for expensive external positioning systems such as laser trackers or motion capture equipment, as well as sensitivity to environmental factors, including map features and landmark availability. To address these limitations, Scheideman et al. [46] developed a cost effective evaluation method using artificial landmarks (AR tags) and onboard vision systems to assess robot positioning accuracy within a 25 square metre indoor environment. The method achieves AR tag tracking accuracy of 1.4 mm ± 1.5 mm within the 0.8 metre sampling range, and can estimate mean absolute trajectory error within an order of magnitude of motion capture ground truth, though confidence decreases for translational errors below 1.4 cm. While this approach offers advantages in terms of affordability and ease of deployment compared to motion capture systems, the method's accuracy remains dependent on visual marker detection precision, camera calibration quality, and environmental lighting conditions, which can introduce measurement uncertainties that affect the reliability of localisation performance assessment.

While research has focused on developing multi sensor fusion systems for mobile robot navigation, evaluating positioning accuracy remains problematic. He et al. [40] developed a system that combined LiDAR, IMU, and wheel odometry data using Gaussian Newton fusion methods and conducted indoor navigation experiments to assess positioning performance. However, their evaluation methodology lacks detail regarding ground truth establishment and accuracy measurement protocols. The authors present "distance error" results without explaining the measurement approach or the validation methods used, which reflects a broader limitation in the current literature: positioning accuracy is discussed without rigorous quantitative assessment frameworks. This highlights the need for standardised evaluation methodologies that provide reliable, comparable accuracy measurements across different mobile platforms and operating conditions.

Early efforts to establish standardised evaluation frameworks in robotics include the Rawseeds benchmarking project, which developed comprehensive datasets and metrics for SLAM, localisation, and mapping algorithms [48]. The Rawseeds approach utilised multi sensor ground truth collection systems, combining vision based tracking (5 camera network) with laser based positioning (4 Sick LMS200 scanners) to establish reference trajectories for algorithm evaluation. While Rawseeds introduced valuable performance metrics such as Self Localisation Error and Integral Trajectory Error, the framework has significant limitations for mobile platform assessment. The system relies on pre-recorded datasets rather than real time evaluation, limiting its applicability to algorithm development rather than practical robot performance assessment. Additionally, the ground truth collection systems covered only small portions of the test environment, constraining the scope of possible evaluation.

Recent advances in simulation based benchmarking have attempted to address some limitations of real world evaluation systems. Spiess et al. developed BLOCSIE, a comprehensive simulation based benchmark for localisation algorithms using Unity game engine with realistic sensor noise modelling [49]. Their approach demonstrated the importance of data driven noise models versus constant standard deviation assumptions, showing significant performance differences across algorithms when realistic sensor characteristics are modelled. However, BLOCSIE remains limited to simulation environments and focuses on algorithm evaluation rather than practical platform performance assessment. While simulation based approaches offer controlled testing conditions and cost advantages, they cannot fully capture the complexities of real world mobile platform operation, including mechanical variations, environmental factors, and platform specific performance characteristics that impact industrial deployment decisions.

Recent benchmarking efforts for indoor mobile robot localisation [50] have established methodologies for evaluating SLAM and navigation algorithms using fiducial marker based ground truth systems. Whilst these approaches successfully assess algorithm performance in changing environments, they assume reliable underlying platform mechanics and do not isolate platform level positioning accuracy from algorithm level localisation performance. Our methodology addresses this gap by directly measuring mechanical platform accuracy independent of localisation algorithm choices, providing complementary data essential for complete mobile manipulator system characterisation.

The frame alignment problem solved here is related to a family of established techniques but is not identical to any of them. Rigid point set registration and iterative closest point alignment assume a single known point set related by one rigid transform; here the reflector mounting offsets are themselves unknown and coupled to the robot's per pose reported orientation, which places the problem closer to robot world (hand eye) calibration and extrinsic sensor calibration than to single set registration. The laser tracker fulfils the ground truth role that motion capture rigs or fiducial marker arrays play in comparable localisation evaluation studies, but at markedly higher point accuracy and without dependence on marker detection or lighting. The present workflow therefore contributes not a new estimator but an application specific, externally referenced evaluation procedure that isolates base positioning accuracy across platforms complementary to the algorithm level benchmarks above, which assume reliable underlying platform mechanics.

Despite significant advances in mobile robot navigation and SLAM technologies, several critical gaps persist in the evaluation of mobile platform positioning accuracy for industrial applications. First, there is no standardised methodology for quantifying and comparing positioning accuracy across different commercial platforms, with existing studies relying heavily on simulation environments or platform specific evaluation protocols that prevent meaningful cross platform comparison. Second, current research lacks comprehensive real world accuracy assessment under controlled conditions, with most studies focusing on navigation capability rather than quantitative positioning precision. Third, existing evaluation approaches either require prohibitively expensive motion capture systems or rely on lower accuracy methods (such as AR tags) that introduce their own measurement uncertainties. Finally, there is insufficient understanding of the relationship between static positioning accuracy and dynamic trajectory following performance, particularly for precision critical applications like NDE inspection.

This study addresses these gaps through three contributions: (1) a standardised experimental workflow using laser-tracker ground truth (6 µm) that is substantially more accurate than fiducial-marker methods and more accessible than a permanently installed motion-capture volume; (2) a comparative characterisation of five commercial platforms under identical conditions; and (3) a coupled calibration and evaluation procedure that accommodates different platform architectures and reports base-positioning accuracy as a design input for NDE deployment.

## 1. Experimental Approach

### 3.1 Platforms and Experimental Setup

Given the absence of standardised testing protocols for mobile platform accuracy assessment, this study develops a novel experimental methodology inspired by [51]. The proposed approach addresses limitations identified in prior studies, though the comprehensive nature of the testing protocol presents implementation challenges.

Following the approach in [15], a Leica Absolute Tracker AT901 serves as the external reference measurement system for ground truth validation. This 3-DOF laser tracker employs coherent laser interferometry to measure position with 6 micrometre accuracy over its operational range. The system's lock on feature maintains continuous tracking of the spherically mounted retroreflector (SMR) during platform motion, with the primary limitation being its 60 degree field of view [52].

As a single target 3 DOF system, the tracker measures three dimensional position only; platform orientation is not measured directly by the tracker and is instead handled as described in Sections 3.2 reconstructed geometrically from the four corner reflectors in the static case, and taken from the robot's reported pose (with the attendant drift dependence) in the single reflector dynamic case.

#### 3.1.1 Platform Selection and Configuration

Five commercially available mobile platforms were selected for comparative analysis: MiR250, KMP1500, Husky UGV, Boston Dynamics Spot, and KUKA KMR. These platforms represent diverse locomotion modalities: differential drive (MiR250, KMP1500, Husky), legged locomotion (Spot), and omnidirectional mecanum wheels (KMR).

To ensure experimental consistency, all platforms were constrained to point turn manoeuvres at trajectory waypoints, effectively disabling strafing capabilities for the Spot and KMR platforms. Additionally, baseline configurations were used without supplementary sensors or performance enhancements—a critical consideration for platforms like Husky and Spot that support extensive additional sensor integration.

#### 3.1.2 Platform Specifications

This is the detailed information about the platform and dimension.

- MiR250: A compact autonomous mobile robot (580 × 800 × 300 mm) , designed for mobile manipulation applications with Universal Robot URseries arm compatibility. The platform integrates front and rear 2D LiDAR scanners, a 3D camera, and ultrasonic sensors for simultaneous localisation and mapping (SLAM) and obstacle detection [24].
- Boston Dynamics Spot: A quadrupedal robot (1100 × 500 mm footprint, 191 mm sitting height) equipped with six forward facing cameras and three cameras on each lateral and posterior face. The platform supports mobile manipulation through the optional Spot Arm. In baseline configuration, navigation relies solely on visual odometry, as LiDAR based mapping requires separate procurement and integration [23].
- KUKA KMR: An integrated mobile manipulator featuring the KUKA iiwa collaborative arm on an omnidirectional base (1080 × 730 × 630 mm) . Mecanum wheels enable translational movement in any 2D direction without reorientation. The platform incorporates dual corner mounted LiDAR scanners with integrated SLAM capabilities [21].
- KUKA KMP1500: The latest generation KUKA mobile platform utilising differential drive locomotion (1300 × 900 × 263 mm) . While currently lacking an integrated manipulator, the platform development will serve as a mobile base for KUKA arm systems. Sensor suite includes dual LiDAR scanners, front and rear cameras, and top/bottom mounted cameras for QR code localisation. The platform supports both QR code based and SLAM based navigation modes; this study focuses exclusively on SLAM performance [20].
- Clearpath Husky UGV: A differential drive platform (990 × 740 × 381 mm) designed for payload integration. In baseline configuration, the platform relies exclusively on wheel odometry for localisation, requiring external sensor integration for environmental perception and mapping capabilities [22].

| Platforms | Length (mm) | Width (mm) | Hight (mm) |
|---|---|---|---|
| MiR250 | 580 | 800 | 300 |
| BD-Spot | 1100 | 500 | 191 (Sitting) |
| KUKA KMR | 1080 | 730 | 630 |
| KUKA KMP | 1300 | 900 | 263 |
| Clearpath Husky | 990 | 740 | 381 |

Table 3.1: Platforms dimensions.

The experimental area was set up with dimensions of 5 metres in width and 10 metres in length. To fully utilise the experimental area and based on the dimensions of the largest robot a 2 metres translation along Y and 5 metres along X was planned. (Fig 3.7) and (Fig 3.8)

Each platform started at a designated position, moved 5 metres along the Y axis, turned 90 degrees, moved 2 metres along the X axis, turned 90 degrees again, and continued this pattern until returning to the starting position. This sequence was repeated five times, ensuring each platform traversed the same path repeatedly. Positional data from the robot were continuously collected during movement, while the laser tracker simultaneously recorded data. A reflector was mounted on each robot to enable the laser tracker to monitor movement. However, due to the reflector’s limited field of view, the robot had to stop at each corner after turning to manually adjust the reflector toward the laser tracker before resuming movement. (Fig 3.6)

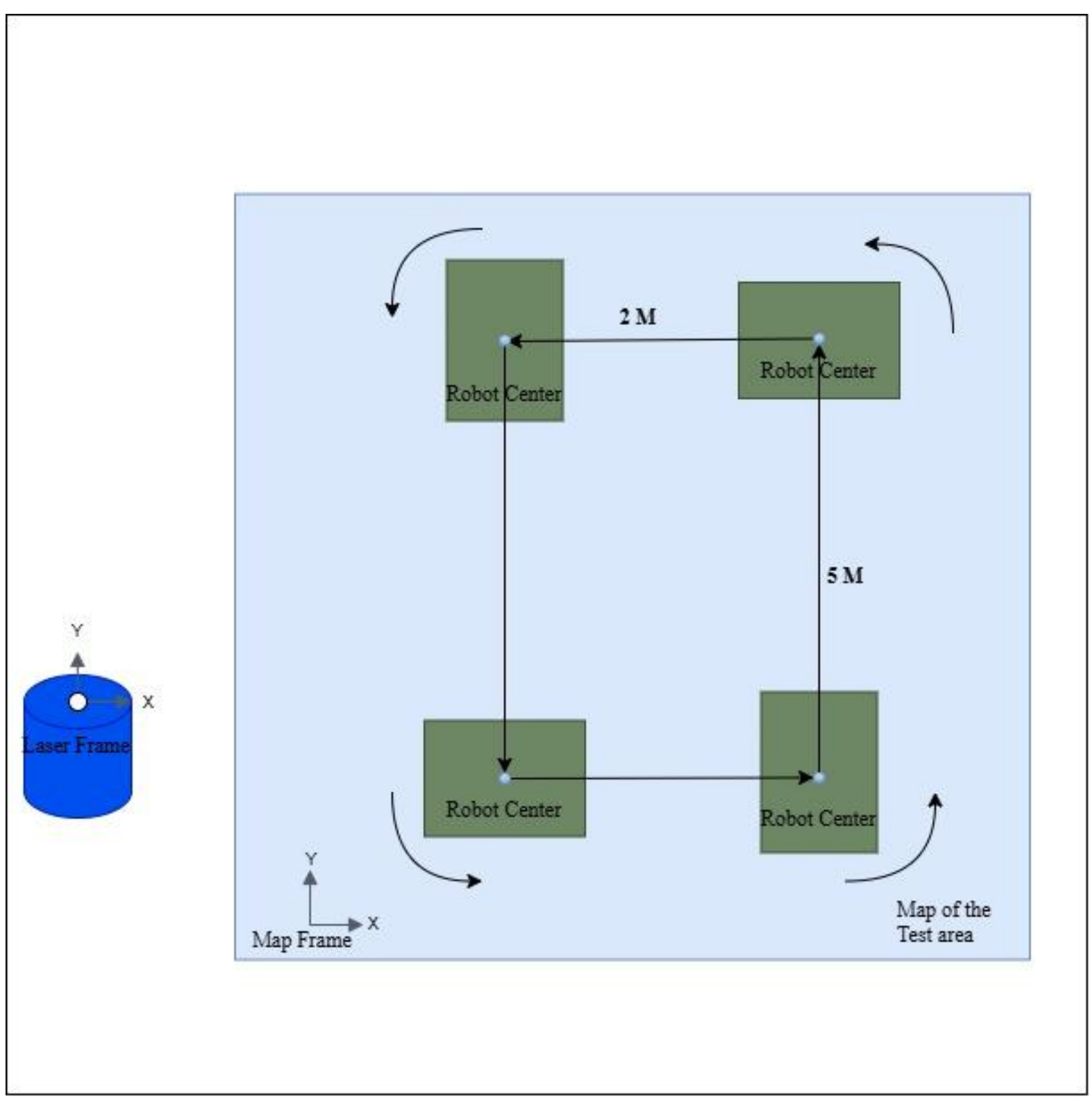

Fig 3.6: Robot path dimensions.

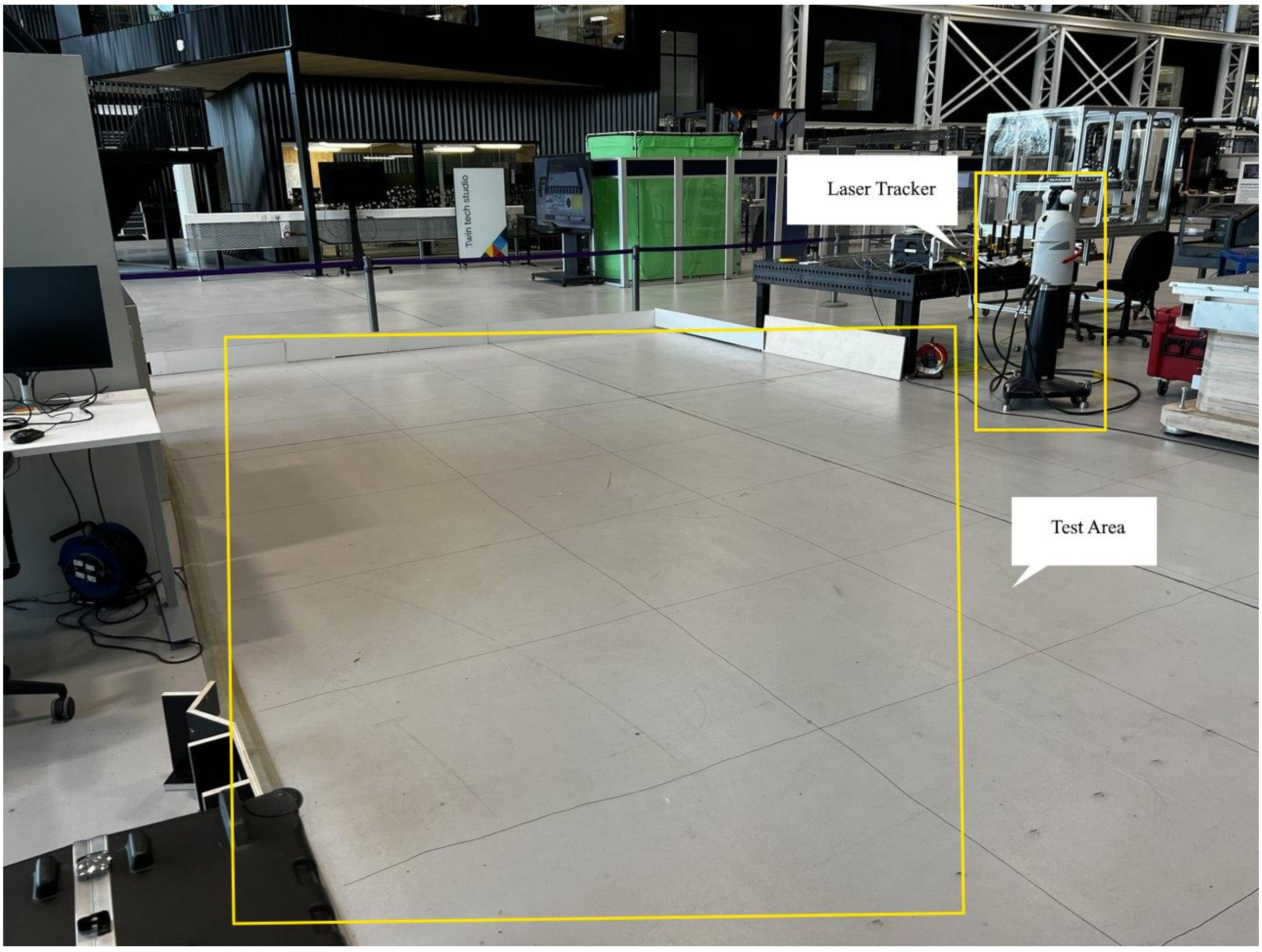

Fig 3.7: Experimental area.

One of the main challenges in this experiment was establishing meaningful accuracy metrics for each platform due to fundamental coordinate frame incompatibilities. Since the robots report their position in their own map reference frame and the laser tracker report the position of the reflector in its own laser frame, there is no direct translation between the two data and it cannot be linked to gather easily. Since these frames are entirely different, comparing data from one system to the other without proper calibration would be meaningless [54].

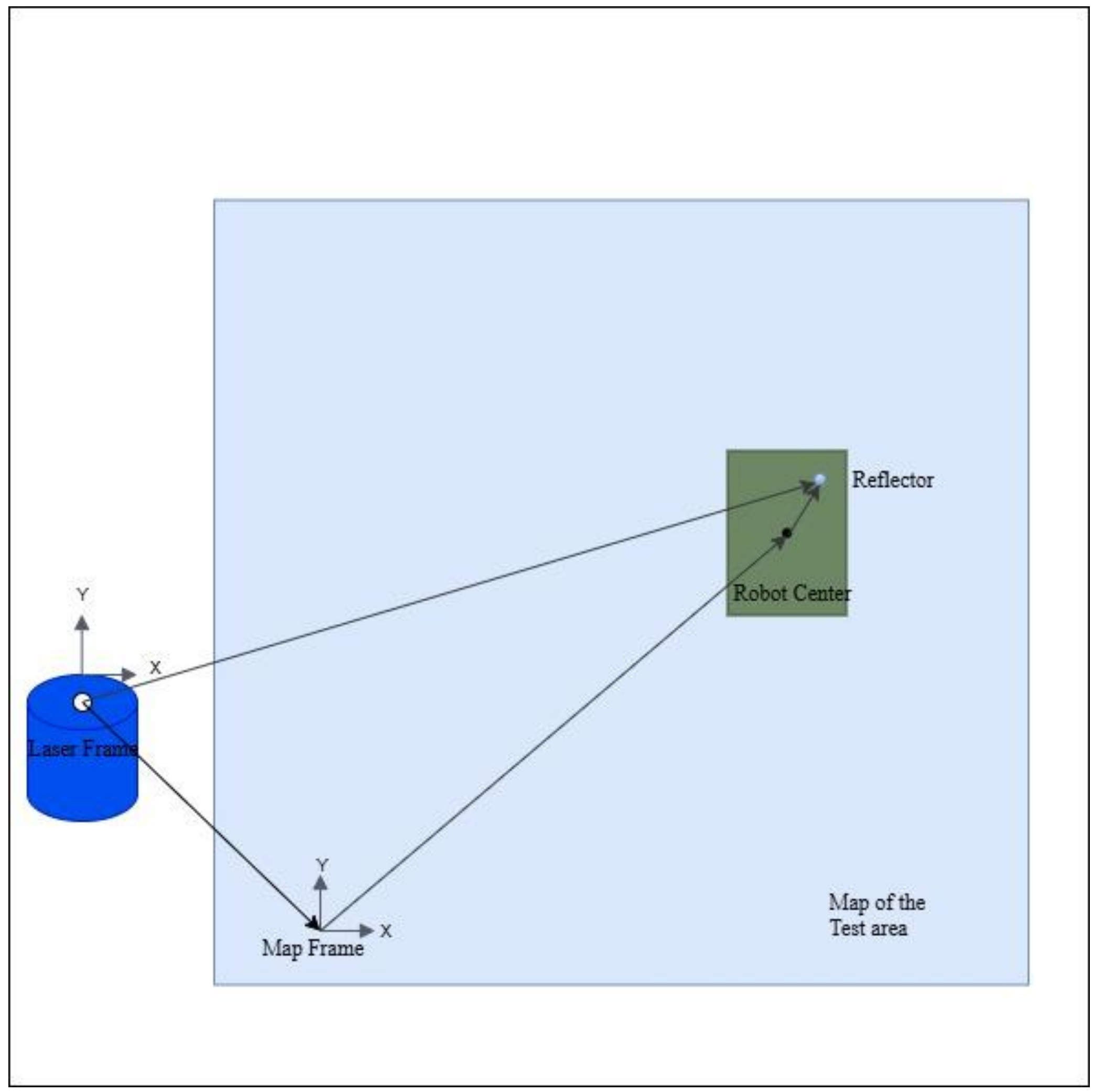

Fig 3.8: Reference frames relation in the test area.

### 3.2 Reference Frame Calibration Methodology

To address this challenge, a reference frame calibration methodology was developed to determine the unknown transformation parameters relating robot reported positions to laser tracker measurements of the physical reflector location. This calibration process must solve for both the spatial relationship between coordinate systems and the physical offset between the robot's reported centre point and the actual reflector position.

Figure (3.7) illustrates the reference frame calibration geometry. The laser tracker measures the reflector position mounted on the robot's corner, while the robot reports its centre position and orientation in its map frame. The calibration must determine: (1) the transformation matrix between laser tracker and robot coordinate systems, and (2) the physical vector from the robot centre to the reflector mounting location.

**Mathematical Framework:**

The transformation between laser tracker frame and robot frame can be expressed as:

$$^{L}P_{reflector} = \, ^{L}R_{R} \cdot \, ^{R}P_{reflector} + \, ^{L}T_{R} \quad (3.1)$$

Where:

$^{R}P_{center} = robot\ center\ position\ in\ robot\ frame$

$^{B}d_{reflector} = reflector\ offset\ vector\ in\ robot\ body\ frame\ (constant)$

$^{R}R_{robot} = 3x3\ rotation\ matrix\ from\ robot's\ orientation$

$^{R}P_{reflector} = position\ of\ reflector\ in\ robot\ frame$

$^{L}P_{reflector} = Position\ of\ Reflector\ in\ Laser\ frame$

$^{L}R_{R} = 3x3\ rotation\ matrix\ from\ robot\ frame\ origin\ to\ laser\ frame\ origin$

$^{L}T_{R} = translation\ vector\ from\ robot\ frame\ origin\ to\ laser\ frame\ origin$

$^{L}P_{reflector} = {}^{L}R_{R} \cdot {}^{R}P_{reflector} + {}^{L}T_{R}$ Which can be expanded in matrix form as (3.2).

$$\begin{bmatrix} ^{L}P_x \\ ^{L}P_y \\ ^{L}P_z \end{bmatrix} = \begin{bmatrix} r_{11} & r_{12} & r_{13} \\ r_{21} & r_{22} & r_{23} \\ r_{31} & r_{32} & r_{33} \end{bmatrix} \begin{bmatrix} ^{R}P_x \\ ^{R}P_y \\ ^{R}P_z \end{bmatrix} + \begin{bmatrix} ^{L}T_x \\ ^{L}T_y \\ ^{L}T_z \end{bmatrix} \quad (3.2)$$

Similarly, the robot's local frame transformation is given by (3.3) and (3.4).

$$^{R}P_{reflector} = {}^{R}R_{robot} \cdot {}^{B}d_{reflector} + {}^{R}P_{center} \quad (3.3)$$

$$\begin{bmatrix} ^{R}P_x \\ ^{R}P_y \\ ^{R}P_z \end{bmatrix} = \begin{bmatrix} a_{11} & a_{12} & a_{13} \\ a_{21} & a_{22} & a_{23} \\ a_{31} & a_{32} & a_{33} \end{bmatrix} \begin{bmatrix} ^{B}d_x \\ ^{B}d_y \\ ^{B}d_z \end{bmatrix} + \begin{bmatrix} ^{B}P_x \\ ^{B}P_y \\ ^{B}P_z \end{bmatrix} \quad (3.4)$$

Combining these transformations yields the complete relationship (3.5) and (3.6).

$$^{L}P_{reflector} = {}^{L}R_{R} \cdot \left( {}^{R}R_{robot} \cdot {}^{B}d_{reflector} + {}^{R}P_{center} \right) + {}^{L}T_{R} \quad (3.5)$$

$$\begin{bmatrix} ^{L}P_x \\ ^{L}P_y \\ ^{L}P_z \end{bmatrix} = \begin{bmatrix} r_{11} & r_{12} & r_{13} \\ r_{21} & r_{22} & r_{23} \\ r_{31} & r_{32} & r_{33} \end{bmatrix} . \left( \begin{bmatrix} a_{11} & a_{12} & a_{13} \\ a_{21} & a_{22} & a_{23} \\ a_{31} & a_{32} & a_{33} \end{bmatrix} \begin{bmatrix} ^{B}d_x \\ ^{B}d_y \\ ^{B}d_z \end{bmatrix} + \begin{bmatrix} ^{B}P_x \\ ^{B}P_y \\ ^{B}P_z \end{bmatrix} \right) + \begin{bmatrix} ^{L}T_x \\ ^{L}T_y \\ ^{L}T_z \end{bmatrix} \quad (3.6)$$

As illustrated in equation (3.5) and (3.6) three unknown terms needed to be calculated to transform data between the robot frame and the laser tracker frame. Here the $R_{LR}$, $^{r}v_{robot}$ and $t_{LR}$ are unknown. Recognising that the robot's data contained significant noise—a central focus of this research—the authors sought to minimise its impact on the calibration process. To achieve this laser tracker data from each corner of the robot was collected and used to solve the transformation equations separately for each corner. This approach provided more accurate laser data with reduced noise, enhancing the reliability of the calibration process (Fig. 3.8).

### 3.3 Platform Control and Software Integration

All platforms were controlled using standardised software interfaces to ensure experimental consistency and enable reproducible methodology across diverse hardware platforms.

### 3.3.1 ROS Based Platform Control

The MiR250, Clearpath Husky, KUKA KMR, and Boston Dynamics Spot platforms were controlled through ROS interfaces, providing standardised communication protocols with consistent message formats for position commands (geometry_msgs/Pose), velocity control (geometry_msgs/Twist), and odometry feedback (nav_msgs/Odometry). All ROS enabled platforms utilised compatible navigation stack implementations, allowing standardised waypoint navigation through the controller framework.

### 3.3.2 KUKA KMP-1500 Control Interface

The KUKA KMP-1500 was controlled through its native KUKA software interface, providing equivalent waypoint based navigation capabilities and comprehensive position/orientation data logging synchronised with the experimental protocol.

### 3.3.3 Interface Standardisation and Synchronisation

To ensure experimental consistency across the different control paradigms (native KUKA interface for the KMP-1500 and KMR; ROS interfaces for the others), experimental commands were translated into platform specific control signals through a common command layer.

During the calibration phase, temporal synchronisation between measurement systems was not required because both robot odometry and laser tracker measured stationary spatial relationships. At each calibration position, the robot remained stationary for several seconds whilst both measurement systems recorded data. During this static period, the robot continuously reported a fixed position in its map reference frame, whilst the laser tracker simultaneously measured the constant position of the reflector mounted on the robot in the laser tracker's reference frame. Since neither measurement changed during the data collection period—both systems were observing the same unchanging spatial configuration—any temporal offset between the measurement timestamps does not affect the geometric relationship being determined. The calibration algorithm determined geometric relationships between coordinate frames based on these paired spatial measurements. Specifically, the algorithm solved for the transformation parameters (rotation matrix R_LR, translation vector t_LR, and four reflector offset vectors v_robot) that mathematically relate the robot's reported position to the laser tracker's measured reflector positions. This transformation is purely geometric in nature and depends only on the spatial arrangement of the coordinate frames, not on when the measurements were recorded. Therefore, the calibration process remained independent of measurement timing within static positioning periods, requiring only that sufficient data be collected whilst the robot maintained each stationary position.

## 3.4 Multi Corner Calibration Approach

The calibration simultaneously recovers the rigid transformation between the laser tracker frame and the robot frame and the physical offsets of the mounted reflectors, using measurements from four corner reflectors per pose. Employing four reflectors rather than one over constrains the geometry at every pose, improves the conditioning of the estimate, and allows the platform's orientation to be reconstructed geometrically from the corner arrangement rather than relying solely on the robot's reported pose.

The contribution of this work is the externally referenced evaluation workflow, not the estimator used within it. The calibration solves a coupled frame registration and offset estimation problem: because the four reflector offset vectors are unknown and coupled to the robot's per pose reported orientation, the problem is structurally closer to robot world / hand eye calibration and extrinsic sensor calibration than to a single rigid point set registration. A closed form Procrustes or Kabsch alignment, or an ICP type correspondence search, does not apply directly, since those assume a single known point set related by one rigid transform. The laser tracker fulfils the ground truth role that a motion capture rig or fiducial marker array plays in comparable localisation evaluation studies, but at substantially higher point accuracy (single-target position to approximately 6 μm).

**Forward model and parameters.** Following the transformation chain of Eqs. (3.1) (3.6), each corner reflector position expressed through the robot's reported centre position and orientation is mapped into the laser tracker frame by the rigid transform (R_LR, t_LR). The estimated parameters are the frame rotation R_LR, parameterised by three intrinsic Euler angles (so a valid rotation is guaranteed by construction and no explicit orthonormality constraint is required); the frame translation t_LR (three components); and the four reflector offset vectors (twelve components) eighteen parameters in total. Each pose provides four reflector observations, i.e. twelve scalar constraints, so approximately two poses already over determine the eighteen unknowns; the number of poses used per platform is therefore governed by desired conditioning and robustness rather than by any minimum solution requirement.

**Objective and estimator.** The calibration minimises the sum of squared three dimensional Euclidean residuals between the predicted and laser measured reflector positions, summed over all poses and all four reflectors, expressed in the laser frame. The objective is non convex in the Euler angles and is minimised using the L-BFGS-B quasi Newton method from a physically motivated initial guess (reflectors at the nominal robot corners; R_LR initialised to the approximate frame yaw). Ordinary least squares over all poses is used as the reported estimator. This choice is deliberate: the frame transformation is a fixed physical relationship that does not depend on how the platform behaved at any individual pose, and as established in Section 3.5  the laser tracker's measurement noise floor is a few micrometres. Poses with large positioning residuals therefore represent genuine platform mispositioning, not measurement error, and are retained. Accuracy is evaluated over all poses; no pose is excluded on the basis of its own residual, as doing so would selectively discard real platform behaviour and bias the reported accuracy optimistically.

**Robustness cross-check.** To confirm that this estimator choice does not affect the result, a robust variant based on Random Sample Consensus (RANSAC)[55] was run alongside the least squares fit for each platform, and the two recovered transformations compared. RANSAC is a well-established robust-estimation technique; it is used here solely as a safeguard against occasional corrupted measurements, not as a novel method or as a source of improved accuracy. On the KMP-1500, KMR and MiR250 the robust and least-squares transformations agreed to within 0.04 - 0.12° in rotation, with matching accuracy and precision figures, confirming that these datasets are free of gross outliers and that ordinary least squares is the appropriate estimator. The one case in which the robust variant proved substantively useful  the detection of a single corrupted pose in the MiR250 dataset is described in Section 3.5[55]

## 3.5 Method Validation and Parameter Sensitivity Analysis

### 3.5.1 Initial Method Validation

Because the workflow had not previously been applied to mobile platform accuracy assessment, its stability was validated before deployment across platforms.. Initial testing was conducted using the MiR250 platform with systematic variation of both position quantities and visit frequencies to investigate their effects on calibration stability.

**Measurement noise floor.** The repeatability of the laser tracker at a fixed pose was characterised from a static hold of consecutive readings of a stationary reflector. The three dimensional standard deviation was 7.0 µm (2.2, 5.6 and 3.6 µm along the three axes; 6.0 µm in the plane of motion), consistent with the tracker's 6 µm single-target specification and approximately three orders of magnitude below the centimetre level platform positioning errors reported in this study. This confirms that the reported accuracy and precision figures reflect platform positioning behaviour rather than metrology noise, and it justifies treating large positioning residuals as genuine platform error rather than measurement outliers.

**Sample-size sensitivity and convergence.** The recovered transformation was tested against dataset size using 10, 20, 40, 100 and 250 measurements. The calibrated parameters were stable across this range: for example, the recovered X-position of the front-right reflector (Fig. 3.9) is essentially unchanged from ten measurements upward, and systematic and random pose selection yielded equivalent transformations. The calibration therefore converges by approximately ten poses, and any sample count above this threshold does not materially affect the result. All platforms in this study were sampled well above this threshold (35–39 poses), with the MiR250 acquired more densely (249 poses) during an extended run assessing sample count sensitivity; sample count is thus not a confounding factor in the cross platform comparison. The parameters likewise showed low sensitivity to the numerical tolerances of the optimisation, with coefficients of variation below 4% for the yaw angle (2.27%) and the reflector X and Y positions (2.09% and 3.70%).

**Blunder detection and handling.** The robust versus least squares comparison introduced in Section 3.5 additionally served as a data quality check. On the MiR250 dataset the two estimators initially disagreed sharply a rotation difference of approximately 36° and a least squares mean error inflated to the order of metres which was traced to a single pose whose recorded reflector coordinates were physically impossible, exceeding the workspace extent by three orders of magnitude. The corruption arose during data handling rather than from the platform or the tracker. The affected pose was removed by its identifier and the remaining 249 poses re analysed; after removal, the robust and least squares transformations agreed to within 0.04°, and the recovered reflector geometry was consistent to better than 1 mm across all poses, confirming that the exclusion was isolated to that single pose. No other platform required any exclusion. This demonstrates that the workflow surfaces corrupted measurements rather than absorbing them silently.

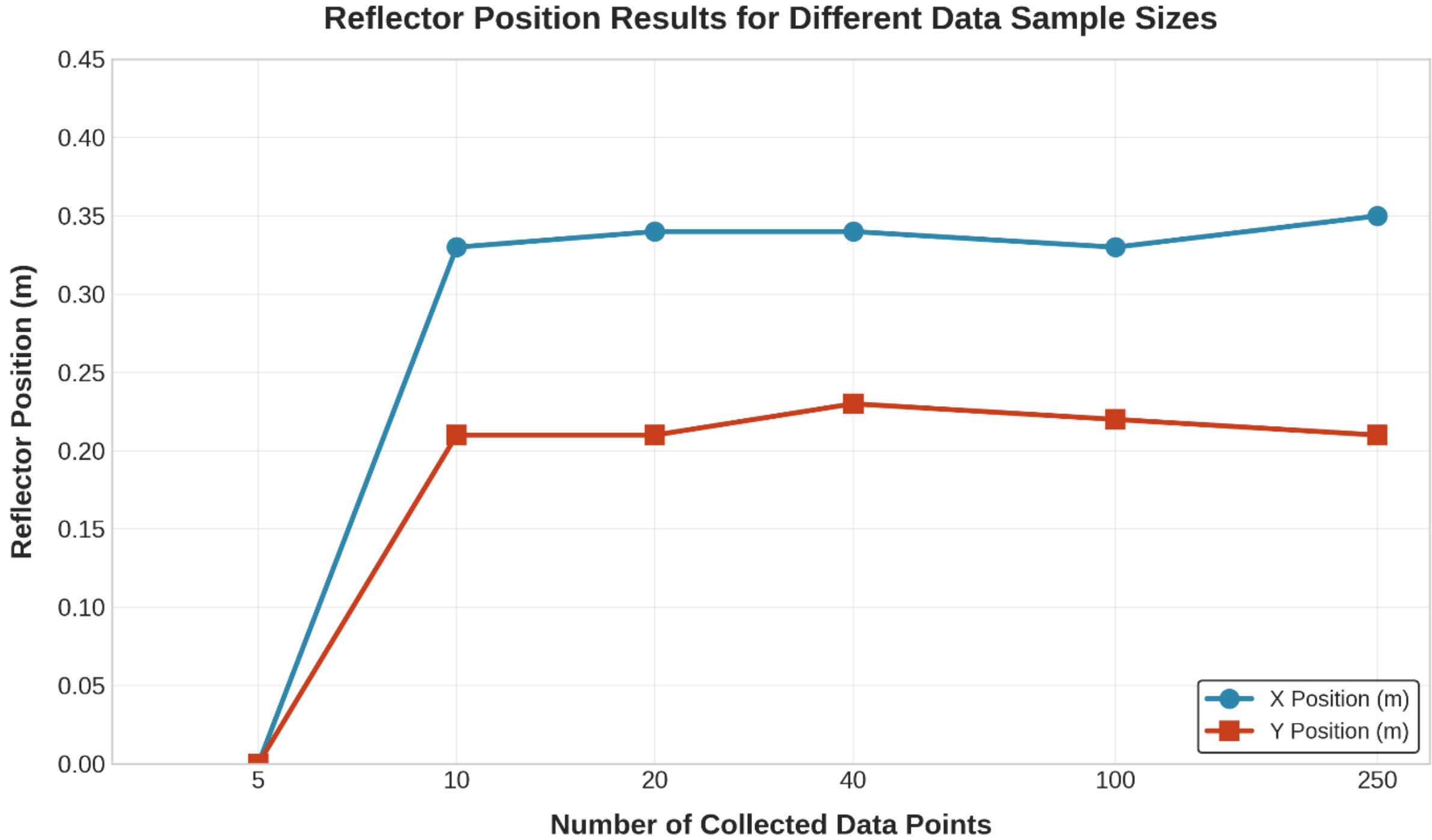

Fig 3.9: Reflector calculated position based on number of data points collected for calibration

### 3.6 Multi Platform Implementation

The calibration and evaluation procedure was applied identically to every platform, so that the resulting accuracy figures are directly comparable despite substantial differences in locomotion, localisation sensing, mapping approach and control interface. Two quantities are reported for each platform and must be kept distinct. *Accuracy* is the mean three dimensional Euclidean error between the robot reported and the laser measured position, computed over all poses. *Precision* is the planar dispersion $\sigma_{2D} = \sqrt{\sigma_x^2 + \sigma_y^2}$ of the signed in plane errors, characterising the repeatability of positioning in the plane of motion.

For each platform, the frame transformation was estimated by ordinary least squares over all poses, and the positioning error evaluated over the same full set of poses. Because the measurement noise floor is a few micrometres (Section 3.5), poses with large residuals reflect genuine platform mispositioning and are retained in the reported statistics; the full error distribution, including its tail, is reported. Where the robust cross check flagged a corrupted sample (Section 3.5, the MiR250), that sample was removed for the stated physical reason before the least squares fit; no other exclusions were applied. Any inlier fraction quoted under the robust cross check is reported only as a diagnostic of data cleanliness, not as a basis for the accuracy figures.

Four of the five platforms could be calibrated. The Clearpath Husky, in its baseline wheel odometry only configuration, could not: no stable transformation converged under any setting, because unbounded odometric drift means the platform's reported poses do not correspond to a fixed rigid frame. The Husky is therefore excluded from all quantitative accuracy comparisons, static and dynamic alike. Although the platform physically executed the trajectories, the absence of a valid frame transformation means its motion cannot be referenced to ground truth, so no accuracy figure is reported for it. This exclusion is itself a result: it marks the lower bound of the localisation capability cascade, at which onboard sensing is insufficient for the platform to be externally characterised.

## 3.7 Dynamic Accuracy Testing Protocol

After completing the calibration between the laser tracker and robot coordinate frame for each platform and determining static accuracy performance, the next phase of the experiment focused on measuring and evaluating the dynamic accuracy of each platform. Due to the absence of clear standardised procedures for dynamic accuracy testing in current literature, a novel methodology was developed, inspired by the approach presented in [12].

### 3.7.1 Trajectory Design and Implementation

The dynamic accuracy assessment protocol consisted of a standardised rectangular trajectory (Figure 3.2) with 5 metre and 2 metre segments connected by 90 degree turns. All platforms operated at 1.0 m/s maximum velocity with different acceleration profiles. Each platform executed this trajectory 10 times to assess positioning consistency and error accumulation over multiple iterations.

The MiR250, Clearpath Husky, KUKA KMR, and KUKA KMP-1500 platforms successfully completed the full 10 iteration testing protocol under identical conditions. However, the Boston Dynamics Spot platform encountered operational constraints due to battery limitations (90 minute average operational duration) and navigation system constraints (absence of integrated LiDAR scanning capabilities requiring manual intervention). Consequently, the Spot platform completed only 6 iterations before battery depletion, as restarting would have required repeating the entire calibration process. Data collection throughout the experiment included continuous position tracking, measurement of actual versus commanded positions at each waypoint, recording of cumulative positional drift over multiple iterations, and assessment of repeatability across trajectory cycles.

### 3.7.2 Temporal data alignment

The dynamic accuracy measurement procedure encountered several significant technical challenges that required innovative solutions. The Leica laser tracker’s reflector has a limited field of view, which caused tracking loss during robot turning manoeuvres at each corner of the rectangular trajectory. To address this limitation, each platform was programmed to stop at each corner waypoint, allowing for manual reflector repositioning and angle adjustment to maintain optimal alignment with the laser tracker before proceeding to the next trajectory segment. Measurements were temporarily suspended during turning manoeuvres and resumed once proper tracking conditions were re established, ensuring data quality while maintaining trajectory integrity.

A more critical technical challenge arose from fundamental differences in time reference systems and data acquisition characteristics between the robot platforms and laser tracker.

The temporal definition of one second interval differs between systems: robot platforms and their control PCs utilise CPU based timing calculations, while the laser tracker employs a liquid crystal oscillator for timing reference [52]. These different timing mechanisms introduce subtle but significant temporal discrepancies between the measurement systems. Furthermore, data collection frequency was not constant in mobile platforms and was heavily influenced by robot hardware performance and computational workload, causing frequency variations during movement. In contrast, the laser tracker maintained a constant, user configurable frequency that provided more stable and accurate timing reference.

Since the laser tracker and ROS systems operate independently without direct communication, a novel synchronisation methodology was developed to address these temporal discrepancies. Start and end times for dynamic motion sequences were manually identified in both datasets, and a velocity profile was generated using robot position data and ROS reported timestamps. Speed variance was calculated across each motion segment and applied consistently to both robot and laser tracker datasets. A velocity threshold of 0.05 m/s was established to identify and match the start and end points of motion sequences in both datasets, ensuring consistent time interval alignment.

To resolve the frequency mismatch between the two data acquisition systems, the robot's data was interpolated to match the number of data points recorded by the laser tracker, as the laser tracker's timing stability made it the more reliable temporal reference. This interpolation approach allowed for direct point to point comparisons between the datasets. However, this interpolation method introduces several limitations that affect the accuracy of dynamic measurements. The interpolation assumes a fixed rate between intervals, which provides accurate results when the robot maintains constant velocity but becomes less reliable during acceleration or deceleration phases. More significantly, the interpolation process introduces positional errors in the results, particularly when robot speeds are higher, as the difference between the actual robot location and the interpolated position increases with velocity. These temporal synchronisation limitations inherently affect the robot dynamic accuracy experiment and must be considered when interpreting the results. This velocity based synchronisation approach, combined with data interpolation, enabled successful alignment of the disparate data streams while acknowledging the measurement uncertainties introduced by the synchronisation process.

### 3.7.3 Methodology Validation

Despite these technical challenges, the implemented solutions maintained experimental rigor and data quality throughout the testing process. The segmented motion protocol, while introducing brief interruptions at corner waypoints, preserved the fundamental characteristics of dynamic accuracy measurement while accommodating hardware limitations inherent in the laser tracking system in regards to the reflector field of view. The velocity based synchronisation methodology provided a robust framework for temporal alignment that could be consistently applied across all platforms, regardless of their specific control architectures or computational capabilities. The resulting dataset, even with the reduced iteration count for the Spot platform, remained valid for comparative analysis as the fundamental measurement methodology and trajectory characteristics were maintained consistently across all platforms. This dynamic testing approach provides a standardised framework for evaluating mobile platform accuracy under realistic operational conditions while accounting for both platform specific limitations and measurement system constraints that may affect industrial deployment scenarios.

# 4 Results

This section presents the results of the mobile platform accuracy experiment, encompassing both static and dynamic accuracy assessments. The results are organised into static accuracy findings derived from the calibration process, followed by dynamic accuracy comparisons across all platforms, and individual platform performance analysis throughout the 10 round dynamic testing protocol.

## 4.1 Static Accuracy Results

Two quantities are reported: *accuracy* (3-D Euclidean error between reported and measured position) and *precision* (planar dispersion $\sigma_2$D of signed in-plane errors). Both median (typical pose) and mean with 95% CI are given, as several distributions are skewed

| Platform | Localisation | n | Median | Mean± 95% CI | | 95th | Max | %>10 mm |
|---|---|---|---|---|---|---|---|---|
| KMP-1500 | LiDAR SLAM | 36 | 8.2 mm | 8.7 [7.7-9.8] | 8.9 | 14.2 | 14.4 | 41.7 |
| MiR250 | LiDAR SLAM | 249 | 24.9 mm | 25.6 [24.0-27.2] | 28.4 | 48.8 | 61.3 | 88.8 |
| KMR | LiDAR SLAM | 35 | 34.1 mm | 39.0 [33.6-45.2] | 42.1 | 74.3 | 90.3 | 100 |
| Spot | Legged Odometry | 39 | 63.5 mm | 77.3[63.8-92.2] | 89.5 | 184.4 | 207.8 | 100 |
| Husky | Wheel Odometry | _ | _ | Uncalibrated | _ | _ | _ | _ |

Table 4.1: Static accuracy error for each platform.

**KMP-1500.** Most accurate: median 8.2 mm (mean 8.7, CI 7.7 - 9.8), $\sigma_2$D 8.9 mm. Right tailed most poses well under a centimetre but 42% exceed 10 mm, so even the best platform cannot be assumed accurate on every placement.

**MiR250.** 24.9 mm median (mean 25.6, CI 24.0 - 27.2); tightest CI owing to the larger sample. Near symmetric and spatially uniform consistently offset by ~2.5 cm rather than usually accurate with occasional errors.

**KMR.** 34.1 mm median (mean 39.0, CI 33.6 - 45.2), 4 – 5 times less accurate than the KMP-1500 despite the same manufacturer, LiDAR SLAM approach and control interface; every pose exceeded 10 mm. The two platforms differ mainly in the generation of their localisation and drive hardware, so this pair isolates that effect while holding vendor and method constant. (KMR uses its KUKA developed software, but KMP has ROS2)

**Spot.** Baseline legged odometry, no map: least accurate calibratable platform, 63.5 mm median (mean 77.3, CI 63.8 - 92.2). Error strongly anisotropic ($\sigma$-x 35, $\sigma$-y 82 mm) and zero mean in both axes genuine positioning spread, not a systematic offset.

**Husky.** Wheel odometry only: uncalibratable (no stable transform converges). Excluded from all comparisons; the lower bound of the cascade.

Laser tracker repeatability at a fixed pose was 7.0 µm 3D (6.0 µm in plane), three orders of magnitude below the platform errors confirming these reflect platform positioning, not metrology noise. Confidence intervals do not overlap between KMP-1500, MiR250 and Spot; the KMR interval overlaps neither MiR250 nor Spot, giving a fully ordered cascade KMP-1500 < MiR250 < KMR < Spot.

### 4.2 Dynamic Accuracy Analysis

**Method and decomposition.** The segmented trajectory (the platform halts at each corner to aim the reflector) was analysed per segment; each segment is aligned in time by cross correlation of the robot and laser speed profiles. Because segments were captured with independent clocks as previously discussed, a residual per-segment temporal offset remains, which projects into the along track direction (error ≈ lag × velocity). Error is therefore decomposed into cross track (perpendicular to motion) and along track (parallel). Cross track is insensitive to this temporal residual and is the primary reported figure; along track is reported as measurement limited.

| Platform | Cross Track Mean/Median | Along Track | Max Speed reached | Log rate | Dynamic Resolution |
|---|---|---|---|---|---|
| KMP-1500 | 6.9/6.2 mm | 7.5 mm | 0.25 m/s | 50 Hz | 5 mm |
| KMR | 17.6/16.1 mm | ~148 mm | 0.47 m/s | 6.3 Hz | 74 mm |
| MiR250 | 18.5/16.2 mm | ~35 mm | 0.65 m/s | 10 Hz | 65 mm |
| Spot | 112.1/105.9 mm | ~97 mm | 0.93 m/s | 9 Hz | 103 mm |

Table 4.2: Cross track is the comparable cross platform metric (resolved for all). Along track is trustworthy only for KMP, bounded/measurement limits for the others.

### 4.3 Platform Performance Comparison

**Resolution.** Dynamic resolution is set by each platform's logging rate and max speed: 5 mm (KMP-1500, 50 Hz, 0.25 m/s), 65 mm (MiR250), 74 mm (KMR) and 103 mm (Spot). This bounds the along track direction, where inter sample motion is unresolved; cross track is resolved far finer because motion perpendicular to travel is negligible between samples.

**Cross track (path following) accuracy** followed the same cascade as the static results: 6.9 mm (KMP-1500), 17.6 mm (KMR), 18.5 mm (MiR250) and 112.1 mm (Spot). The KMP-1500, with 50 Hz logging and the lowest cruise speed, had negligible temporal residual, its along track (7.5 mm) matches its cross track, so its total dynamic error (11.3 mm) is fully resolved and closely matches its static accuracy (8.7 mm). For the slower logging platforms the along track was dominated by the temporal residual and is reported as bounded. Spot is the exception in the other direction: its cross track (112.1 mm) exceeds its along track, indicating genuine path following drift rather than a measurement artefact, consistent with map free legged odometry with high segment to segment variability (110 ± 66 mm).

**Configuration and kinematic caveats.** Platforms differ in locomotion, localisation, mapping and control interface, and were operated within their own safe envelopes (below 1 m/s) under their native accelerations. Achieved speed, acceleration transients and logging rate therefore differ across platforms. Dynamic figures characterise each platform as configured and safely deployed, not under matched kinematics, and are not an intrinsic mechanical ranking. Cross track is the metric comparable across platforms (resolved for all), along track is trustworthy only for the KMP-1500.

### 4.4 Trajectory Analysis and Path Tracking Performance

The dynamic path following performance of each platform is presented in three complementary forms: the along track and cross track error versus path distance for each platform (Fig: 4.1- 4.4), the distribution of cross track error for each platform (Fig: 4.5-4.8), and the three-dimensional trajectories across all testing cycles (Figs 4.9–4.12).

As introduced in the resolution analysis above, the per-sample error is decomposed into a cross-track component (perpendicular to the direction of travel) and an along-track component (parallel to it). This decomposition is essential to interpreting the dynamic results, because the residual temporal offset between the independently captured robot and laser segments projects almost entirely into the along-track direction (error ≈ lag × velocity), whereas the cross-track component is insensitive to it.

The error versus distance figure makes this distinction directly visible. For the KMR (Fig:4.1), the along track error forms flat plateaus of approximately 200 - 300 mm that reverse in sign between opposing straight segments, producing a regular square wave pattern locked to the trajectory. This forms a constant offset within each segment, changing sign with travel direction, is the signature of a temporal alignment residual, not of platform mispositioning: a genuine positioning error would vary continuously along the path rather than holding a constant value that flips at each corner. The along track error is therefore reported as a measurement limited quantity bounded by the dynamic resolution, not as platform accuracy. The cross track error (blue) remains small throughout, between approximately 0 and 50 mm, and represents the platform's actual path following performance. The error minima at the segment boundaries coincide with the near stationary reflector adjustment periods, where velocity and hence the along track residual approaches zero.

The distribution of the KMR cross track error (Fig: 4.5) is unimodal with a mild positive tail, centred on a median of 16.1 mm (mean 17.6 mm), with the great majority of samples below 30 mm.

The same pattern holds across the slower logging platforms (MiR250 and Spot): a large, plateau form along track error dominated by the temporal residual, and a smaller cross track error that reflects genuine path following. The KMP-1500 is the exception and confirms the interpretation: logging at 50 Hz and driven at the lowest cruise speed, it has a negligible temporal residual, so its along track and cross track errors are comparable and both small (Fig: 4.2), consistent with its fully resolved total dynamic error of 11.3 mm and its close agreement with its static accuracy of 8.7 mm. The measurement artefact is thus platform specific, a consequence of logging rate and cruise speed and not a blanket limitation of the method.

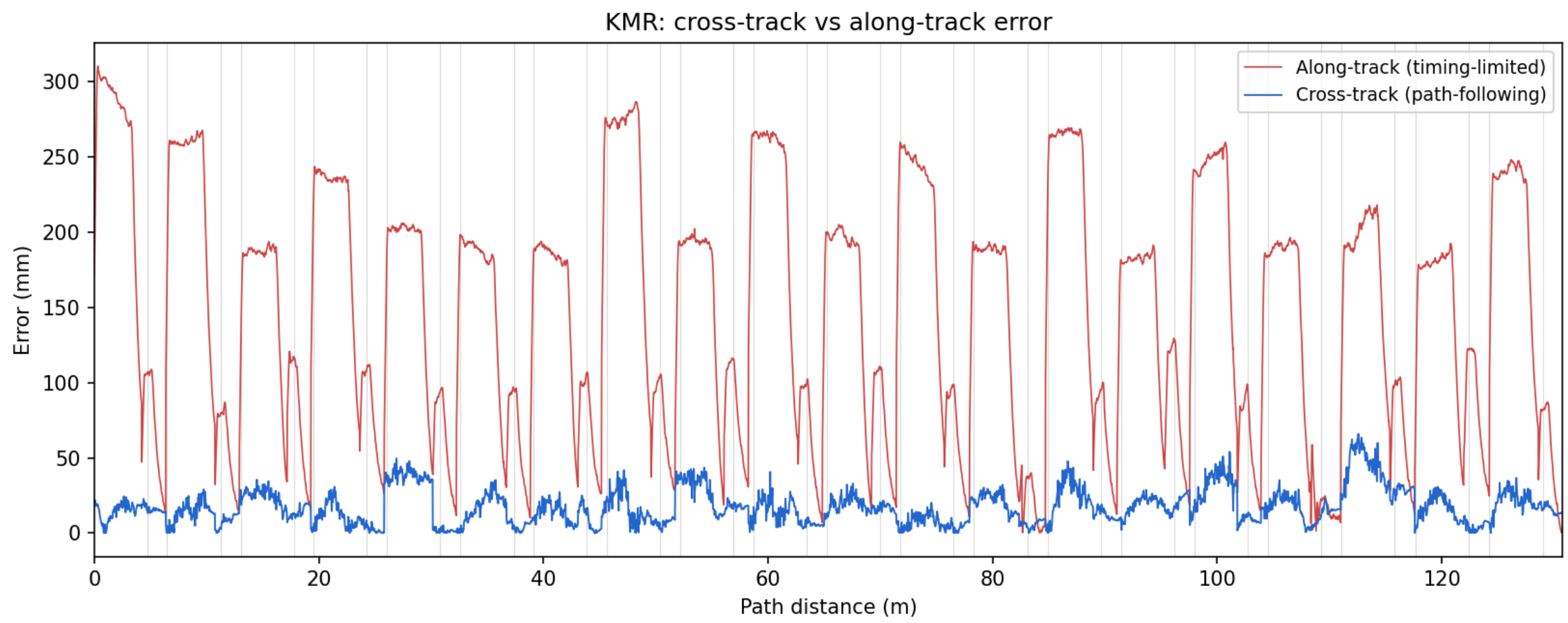

Fig 4.1: KMR Cross track vs Along track Error.

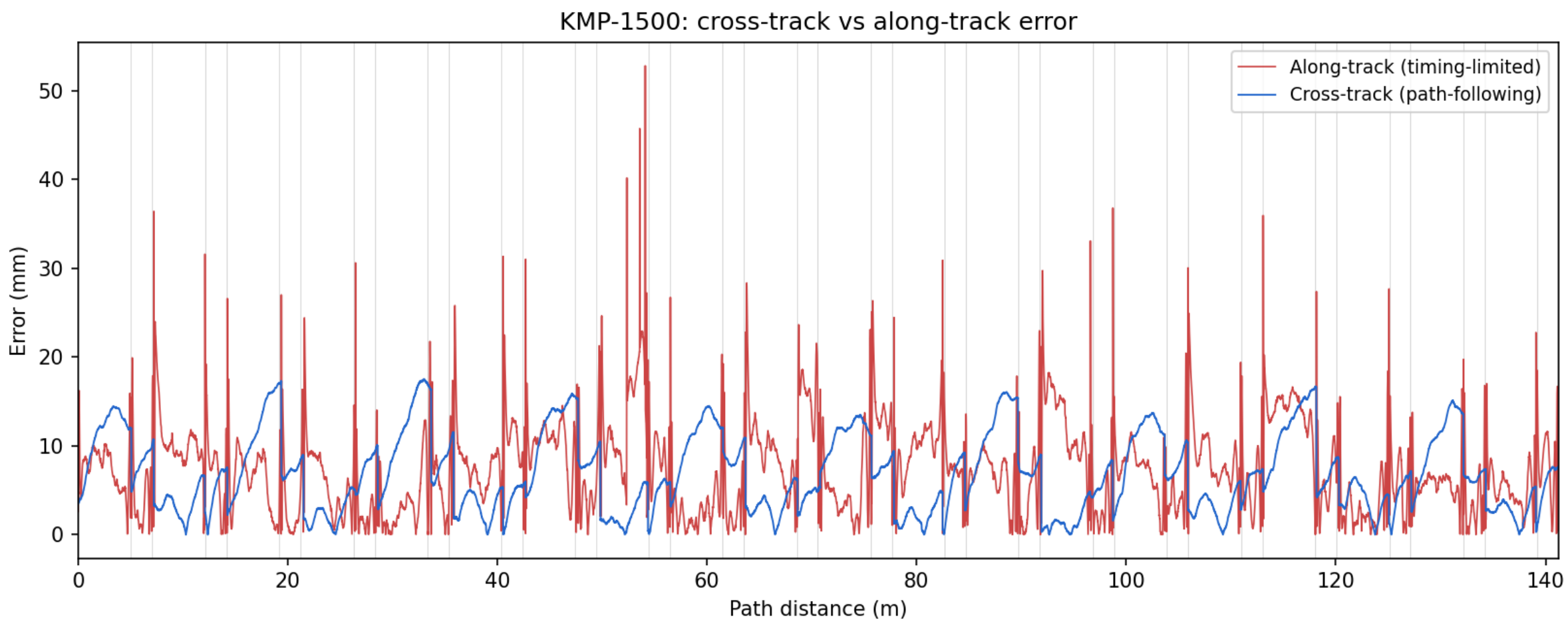

Fig 4.2: KMP-1500 Cross track vs Along track Error

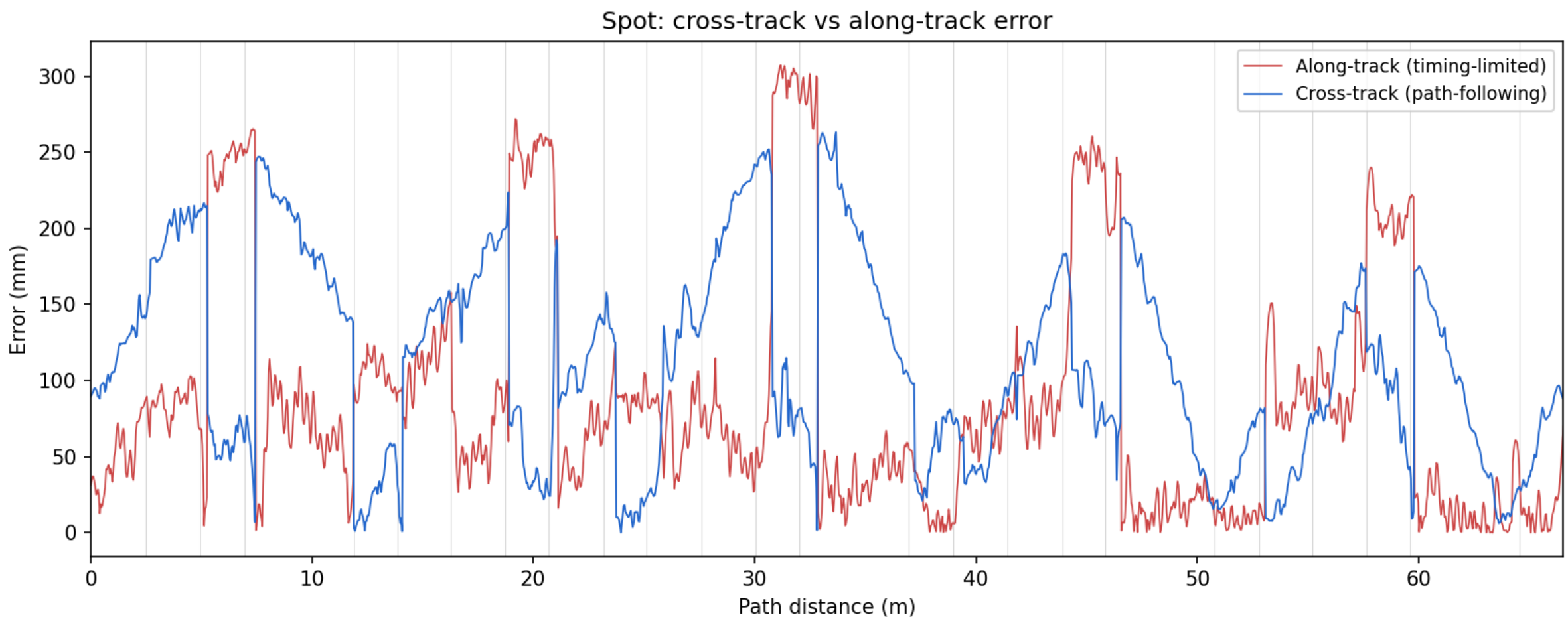

Fig 4.3: Spot Cross track vs Along track Error

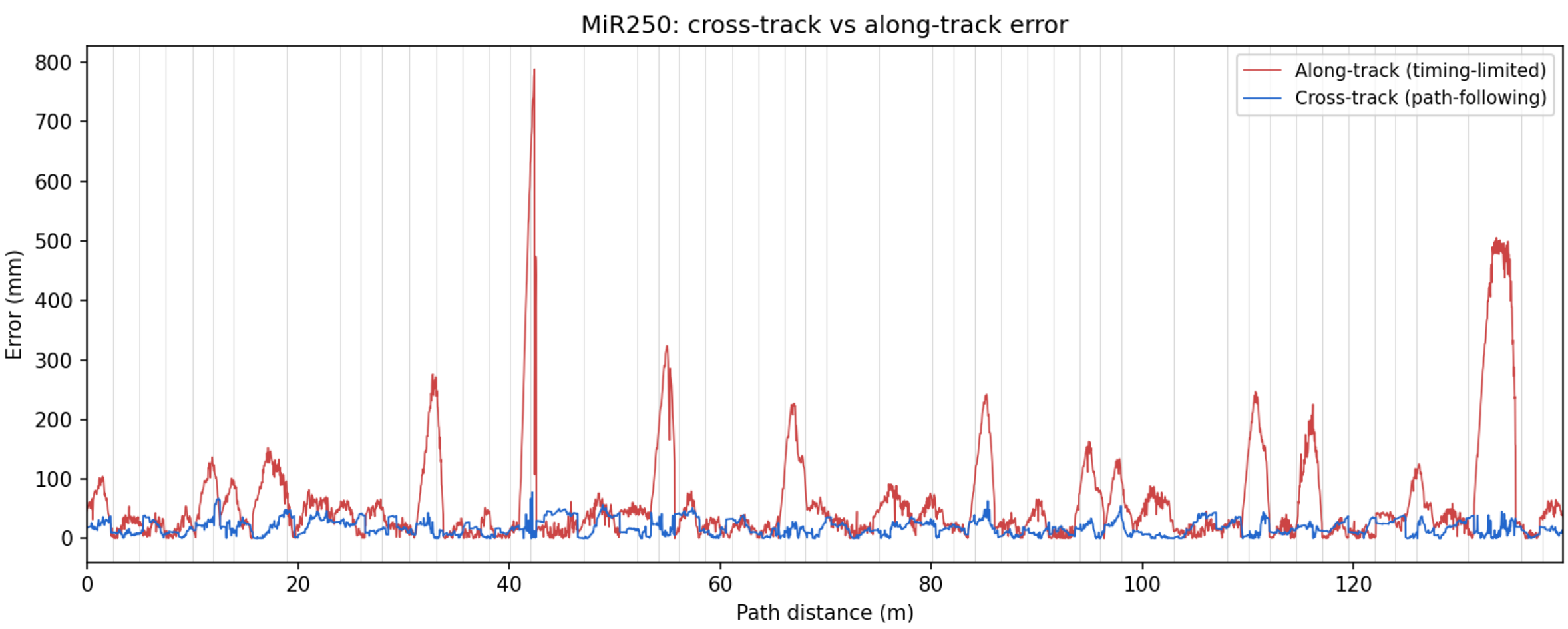

Fig 4.4: MiR250 Cross track vs Along track Error

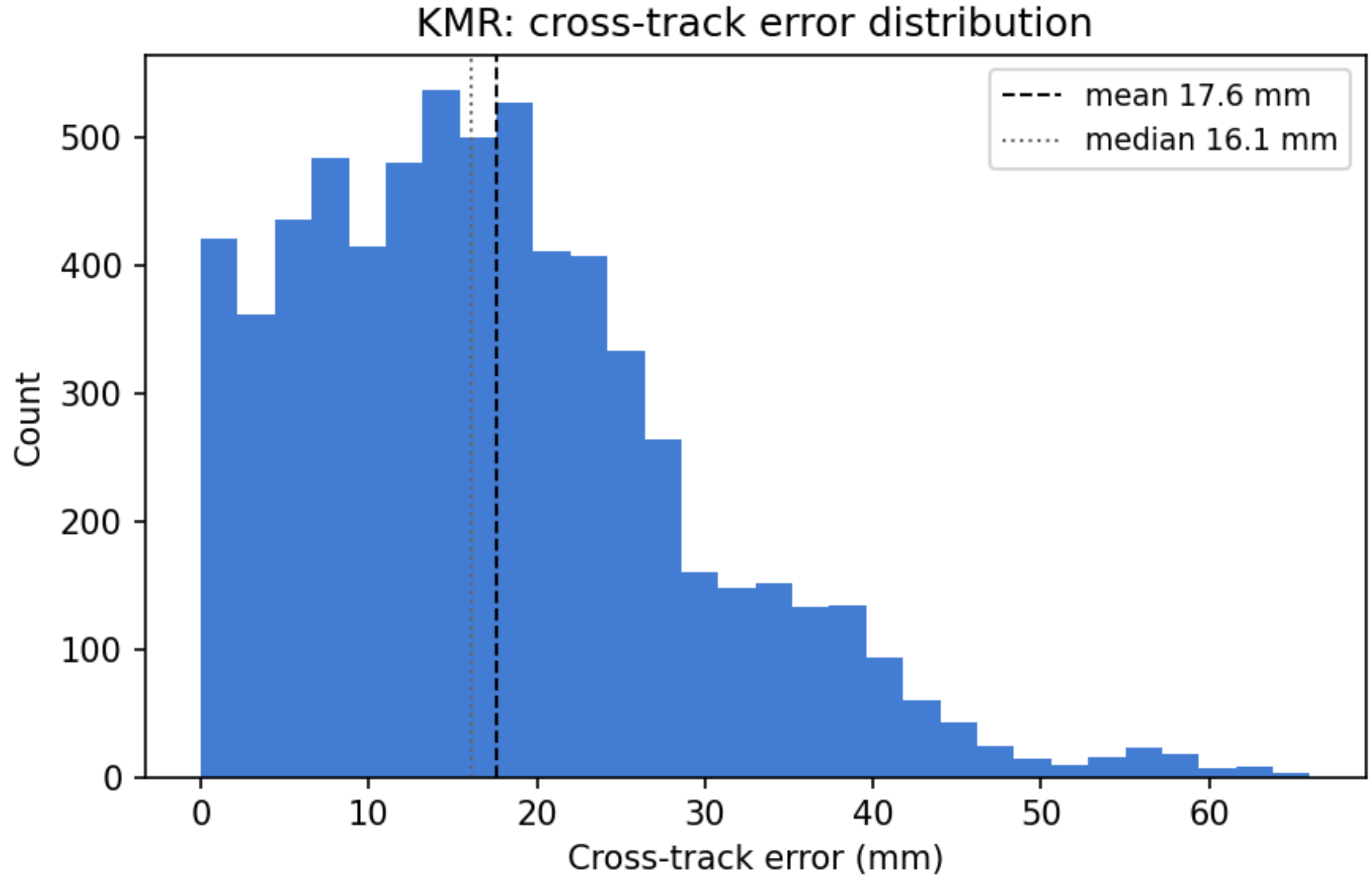

Fig 4.5: KMR Cross track Error distribution

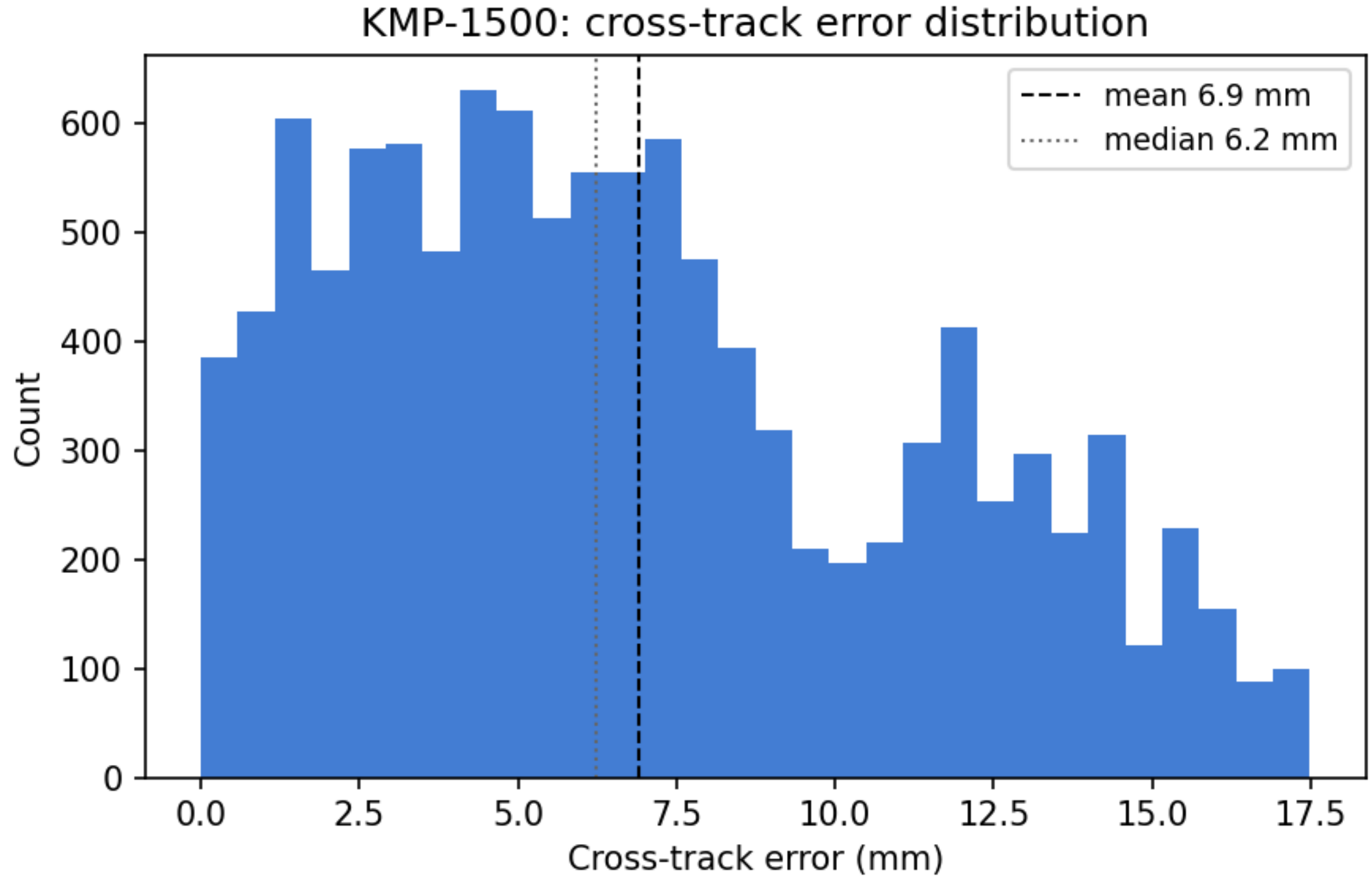

Fig 4.6: KMP-1500 Cross track Error distribution

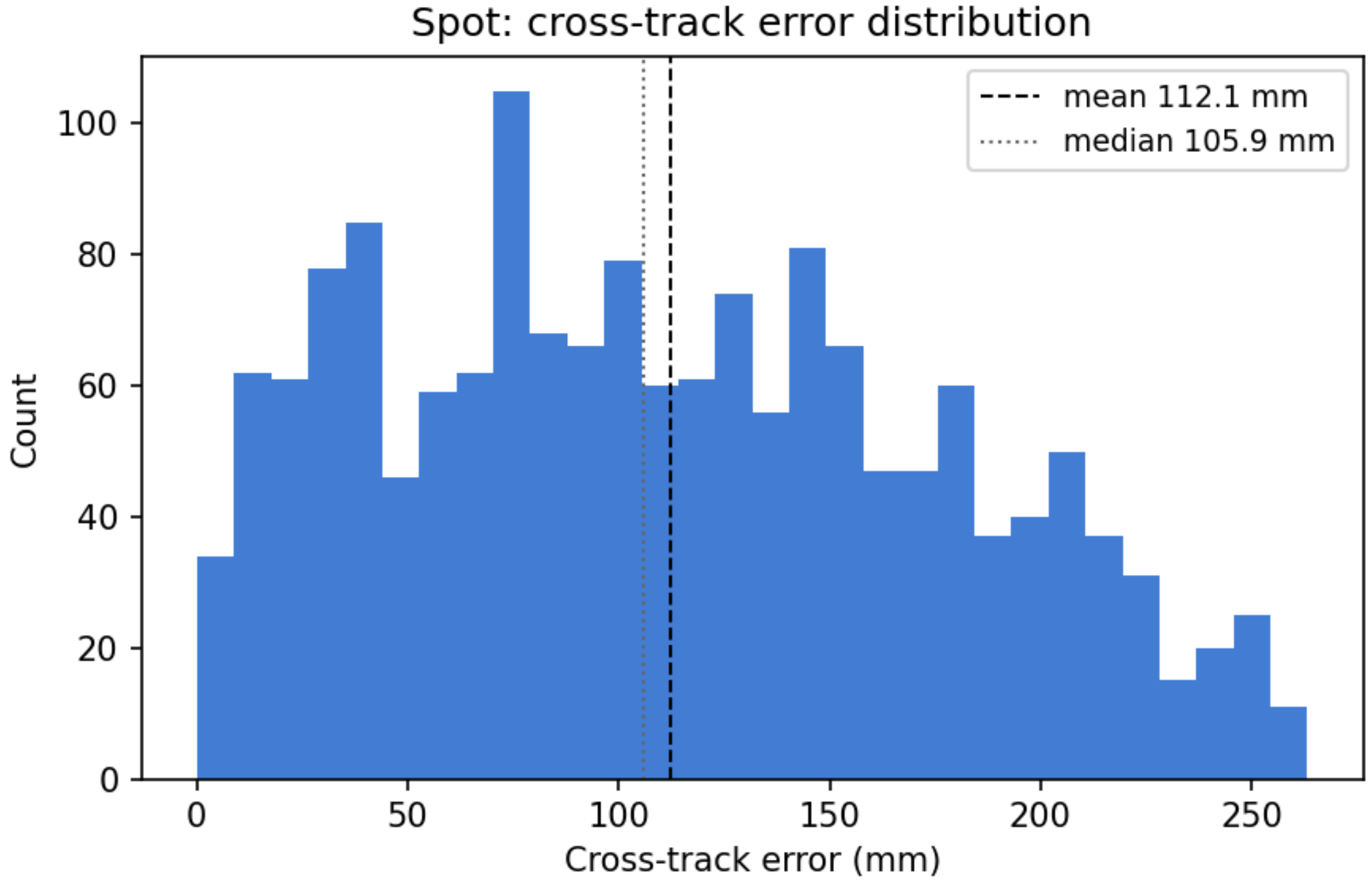

Fig 4.7: Spot Cross track Error distribution

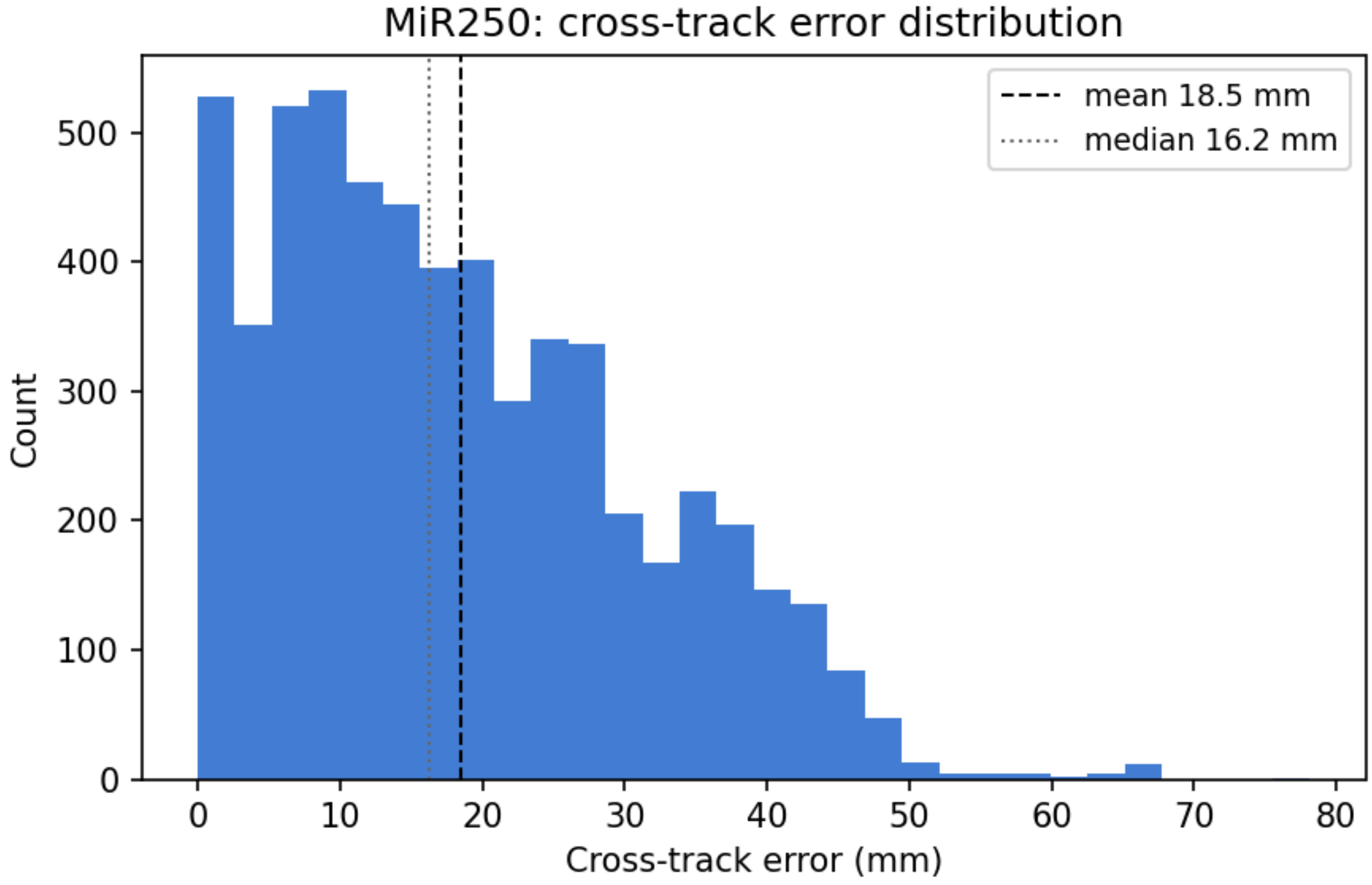

Fig 4.8: MiR250 Cross track Error distribution

### 4.5 Cross-Platform Performance and Industrial Implications

Both static accuracy and dynamic cross track accuracy broadly follow the same localisation capability cascade: the KMP-1500 is most accurate on both measures, the Spot least accurate, with the KMR and MiR250 intermediate. The ordering is not identical between the two, however statically the MiR250 (25.6 mm) is more accurate than the KMR (39.0 mm), whereas in dynamic cross track the two are comparable (18.5 mm and 17.6 mm respectively). Static accuracy is therefore an indicator of, but not a precise predictor of, dynamic path following performance; platform specific factors such as motion control behaviour and localisation update rate influence how static accuracy translates into motion. For this reason no quantitative static to dynamic correlation is claimed.

The practical implication is that positioning accuracy, and indeed whether a platform can be externally characterised at all, tracks its localisation capability rather than its class or manufacturer. The KMP-1500 and KMR, which share manufacturer, LiDAR SLAM sensing and control interface, differ in accuracy by a factor of four to five, attributable to the generation of their localisation and drive hardware; and the Husky, in its wheel odometry only configuration, could not be calibrated at all. Selection for a precision application therefore cannot rely on manufacturer specifications, which do not predict deployed positioning accuracy, but requires the kind of externally referenced characterisation demonstrated here. Among the platforms tested, applications requiring the smallest base positioning error are best served by the newest industrial wheeled platform (KMP-1500), while the more mobile or legged platforms trade positioning accuracy for their respective advantages in manoeuvrability and terrain capability.

The substantial performance variation across platforms underlines the value of a comprehensive, externally referenced accuracy assessment before deployment in precision critical tasks, and the dynamic protocol, through the cross track decomposition isolates genuine path following performance from the measurement resolution, revealing platform differences that a single aggregate error figure would obscure.

## 5 Discussion

### 5.1 Platform Performance Analysis and Industrial Implications

Positioning accuracy tracked localisation capability across the platforms, in both the static (8.7 → 77.3 mm) and dynamic cross track (6.9 → 112.1 mm) results, in the same order. The KMP-1500/KMR pair is especially informative: same vendor, sensing approach and control interface, yet a 4 – 5 times accuracy difference attributable to the generation of the localisation and drive hardware. Accuracy cannot be inferred from a platform's class or manufacturer; it must be measured for the specific unit and configuration deployed.
**Comparison with Fixed Systems and NDE Application:** Fixed manipulators achieve 0.01–0.2 mm [56], [57]. The mobile platforms are roughly two to four orders of magnitude coarser the price of mobility, not a defect. Relative to the ±0.2 - 1.0 mm NDE tolerance.

[58], [59]. the best platform (8.7 mm static) is about one order of magnitude short of the loosest bound and the worst (Spot) nearly two. Precision inspection is performed with the probe stationary over the scan point, so the relevant quantity is static accuracy there the strongest, most repeatable result not dynamic behaviour between points. The measured base accuracy sizes the supplementary sensing each platform needs: closing ~one order of magnitude for the KMP-1500, nearly two for Spot. These describe base placement only; final probe accuracy would compound base error with manipulator and end effector calibration, so the base accuracy is a design input to, and lower bound on, that total.

The platforms also differ in error character: the KMP-1500 is usually excellent but right tailed, while the MiR250 is consistently offset with little tail, implying per placement verification for the former and a uniform correction for the latter.

### 5.2 Cost Performance Trade offs

Fixed high precision manipulators require £200,000-500,000 investments plus infrastructure costs, while mobile platforms range from £50,000-150,000. However, augmenting mobile platforms with necessary metrology systems—laser trackers (£80,000-150,000), vision systems (£20,000-50,000), or force controlled end effectors (£30,000-60,000)—can approach fixed installation costs. Mobile platform deployment is most cost effective when: (1) accuracy requirements align with native centimetre level capabilities, or (2) flexibility benefits justify additional sensing infrastructure costs.

## 6 Conclusions

Across the four calibratable platforms, static base accuracy ranged from 8.7 mm (KMP-1500) to 77.3 mm (Spot), and dynamic cross track from 6.9 mm to 112.1 mm, two to four orders of magnitude coarser than fixed manipulators and one to two orders short of aerospace NDE tolerance. Accuracy and calibratability tracked localisation capability, from the newest LiDAR SLAM platform to the odometry only Husky, which could not be characterised at all. No configuration meets NDE tolerances from the base alone; the contribution is a reproducible, externally referenced workflow that quantifies how far each platform falls short and thus how much supplementary sensing its deployment requires.

**Achievement of Research Objectives:**

This work set out to establish a standardised, externally referenced method for quantifying the base positioning accuracy of commercial mobile platforms for NDE deployment, and to characterise and compare five such platforms under a common protocol. Both objectives were met. A laser tracker based workflow with approximately 6 µm ground truth accuracy was developed and applied identically across platforms differing in locomotion, localisation and control architecture, yielding directly comparable static and dynamic accuracy figures. Four of the five platforms were successfully calibrated and characterised; the fifth (Husky), being uncalibratable in its baseline configuration, itself defined the lower bound of the measurable range. The framework quantified not only each platform's accuracy but the resolution and confounds of the measurement process itself, so that the reported figures are bounded and reproducible rather than nominal.

**Key Findings:**

- Static base accuracy ranged from a median of 8.2 mm (KMP-1500) to 63.5 mm (Spot), with dynamic cross track path following accuracy from 6.9 mm to 112.1 mm, both one to two orders of magnitude short of the ±0.2–1.0 mm aerospace NDE tolerance.
- Accuracy, and calibratability itself, tracked localisation capability across the platforms, from the newest LiDAR SLAM system to map free legged odometry to uncalibratable wheel odometry a consistent cascade in both the static and dynamic results.
- The KMP-1500 and KMR, sharing manufacturer, sensing approach and control interface, differed in accuracy by a factor of 4 - 5, indicating that positioning accuracy is governed by the generation and capability of the localisation and drive hardware rather than by platform class.
- Platforms differed in error character as well as magnitude the KMP-1500 usually accurate but occasionally beyond a centimetre, the MiR250 consistently offset, implying different compensation strategies for each.
- Dynamic accuracy was limited by the measurement resolution set by each platform's logging rate and cruise speed, and by temporal alignment across independently captured segments; cross track error, being insensitive to this, is reported as the reliable dynamic metric.

.

**Significance:**

The significance of this work lies in providing a reproducible, externally referenced basis for selecting and deploying mobile platforms in precision NDE, rather than a demonstration that any current platform meets NDE tolerances. By quantifying each platform's base accuracy and its variability, the framework converts positioning performance into a concrete design input: it sizes the supplementary sensing, external metrology, vision registration or compliant end effectors, that each platform would require to close the residual gap to NDE tolerance, from roughly one order of magnitude for the most accurate platform to nearly two for the least. In doing so it offers researchers and integrators a standardised evaluation procedure and a comparative dataset where previously platform selection decisions rested on manufacturer specifications that, as this study shows, do not predict deployed positioning accuracy.

# 7 Future Research Directions

**Temporal Synchronisation:** Hardware level solutions including Network Time Protocol implementation, shared hardware triggers, or real time kinematic positioning systems to achieve microsecond level accuracy.

**Standardisation:** Uniform SLAM algorithms and navigation software across platforms to eliminate software induced performance variations and enable true hardware comparison.

**Comprehensive Characterisation:** Systematic investigation of velocity accuracy relationships across full operational speed ranges, environmental robustness testing, long term drift analysis, and payload effects from integrated manipulator arms.

**Advanced Metrology Integration:** Photogrammetry systems or multiple laser tracker arrays to overcome field of view limitations while maintaining measurement precision.

**Real time Monitoring:** Development of adaptive control strategies that adjust operational parameters based on detected performance degradation.